\documentclass{article}

\usepackage[preprint]{neurips_2025}

\usepackage[utf8]{inputenc} % allow utf-8 input
\usepackage[T1]{fontenc}    % use 8-bit T1 fonts
\usepackage{hyperref}       % hyperlinks
\usepackage{url}            % simple URL typesetting
\usepackage{booktabs}       % professional-quality tables
\usepackage{amsfonts}       % blackboard math symbols
\usepackage{nicefrac}       % compact symbols for 1/2, etc.
\usepackage{microtype}      % microtypography
\usepackage{xcolor}         % colors

\usepackage{makecell}
\usepackage{multirow, multicol}
\usepackage{xcolor}
\usepackage{booktabs}
\usepackage{enumitem}
\usepackage{graphicx}
\usepackage{colortbl}
\usepackage{wrapfig}

\usepackage{times}
\usepackage{soul}
\usepackage{url}
\usepackage[utf8]{inputenc}
\usepackage[small]{caption}
\usepackage{amsmath}
\usepackage{amsthm}
\usepackage{amssymb}
\usepackage{booktabs}
\usepackage{algorithm}
\usepackage{algorithmic}
\usepackage{enumitem}
\setlist{nosep}
\setlist[itemize]{leftmargin=20pt}

\newtheorem{definition}{Definition}
\newtheorem{claim}{Claim}
\newtheorem{assumption}{Assumption}
\newtheorem{proposition}{Proposition}

\title{Fairness-aware Anomaly Detection via Fair Projection}

\author{Feng Xiao$^{1}$ Xiaoying Tang $^{1,3,4}$  Jicong Fan $^{1, 2}$\thanks{Corresponding author}\\
  $^{1}$The Chinese University of Hong Kong, Shenzhen, China\\ 
  $^{2}$Shenzhen Research Institute of Big Data, Shenzhen, China\\
  $^{3}$The Shenzhen Institute of Artificial Intelligence and Robotics for Society\\
  $^{4}$The Guangdong Provincial Key Laboratory of Future Networks of Intelligence\\
}

\begin{document}

\maketitle

\begin{abstract}
Unsupervised anomaly detection is a critical task in many high-social-impact applications such as finance, healthcare, social media, and cybersecurity, where demographics involving age, gender, race, disease, etc, are used frequently. In these scenarios, possible bias from anomaly detection systems can lead to unfair treatment for different groups and even exacerbate social bias. In this work, first, we thoroughly analyze the feasibility and necessary assumptions for ensuring group fairness in unsupervised anomaly detection. Second, we propose a novel fairness-aware anomaly detection method FairAD. From the normal training data, FairAD learns a projection to map data of different demographic groups to a common target distribution that is simple and compact, and hence provides a reliable base to estimate the density of the data. The density can be directly used to identify anomalies while the common target distribution ensures fairness between different groups. Furthermore, we propose a threshold-free fairness metric that provides a global view for model's fairness, eliminating dependence on manual threshold selection. Experiments on real-world benchmarks demonstrate that our method achieves an improved trade-off between detection accuracy and fairness under both balanced and skewed data across different groups.
\end{abstract}

\section{Introduction}
As machine learning techniques are increasingly being applied in high-social-impact fields such as finance and justice, the fairness of machine learning systems receives a surge of attention. There are growing studies~\citep{larson2016we,hendricks2018women,dastin2022amazon} that exhibit discrimination in real-world machine learning systems. For instance, analysis~\citep{larson2016we} on the COMPAS (Correctional Offender Management Profiling for Alternative Sanctions), a recidivism risk prediction system, shows a strong correlation between recidivism prediction and race, where African-American individuals have a much higher risk of recidivism than Caucasian. It is vital to eliminate or mitigate the possible disparity of machine learning algorithms between different demographic groups to ensure social fairness.

In response, many researchers~\citep{Dwork_Hardt_Pitassi_Reingold_Zemel_2012,zemel2013learning,tzeng2014deep,Hardt_Price_Srebro_2016,Zafar_Valera_Gomez_Rodriguez_Gummadi_2017,agarwal2019fair,oh2022learning,jovanovic2023fare} have begun to propose fairness-aware machine learning algorithms, where one of the most common paradigms is to learn a fair representation depending on a formal fairness principle~\citep{Hajian_Bonchi_Castillo_2016,Hardt_Price_Srebro_2016,zafar2017parity,Gajane_2017}. However, previous works on algorithm fairness often focused on supervised machine learning tasks \citep{Hardt_Price_Srebro_2016,agarwal2018reductions,agarwal2019fair} and the study on the fairness issue in unsupervised anomaly detection~\citep{pang2021deep} is scarce. Anomaly detection (AD), aiming at identifying anomalous samples in data, plays a crucial role in many important fields such as finance, healthcare, social media, and cybersecurity, where demographics, like age, gender, race, ethnicity, and disease, are used frequently. Given multiple groups ($\geq 2$) partitioned by a protected social variable with multiple attribute values, a fair AD method is supposed to ensure equal probabilities of samples being detected as anomalous (or normal) across different demographic groups\footnote{In this work, we focus on group fairness rather than individual fairness.}. However, as shown by \citep{zhang2021towards}, existing AD methods suffer from unfairness to some extent. In addition, \cite{meissen2023predictable} and \cite{wu2024fair} found that the fairness of unsupervised anomaly detection models is easily affected by sample proportion across different demographic groups.

Surprisingly, studies in the literature focusing on fairness in unsupervised AD are quite limited and incomplete. Although there have been several attempts such as \citep{deepak2020fair,zhang2021towards,Shekhar_Shah_Akoglu_2021,han2023achieving}, we still encounter, at least, the following four problems or difficulties:
\begin{enumerate}[label=\Roman*] 
    \item The feasibility and necessary assumptions of ensuring fairness in unsupervised anomaly detection haven't been clearly discussed in the literature.\label{diffi_1}
    \item The trade-off between fairness and utility criteria (such as accuracy) does not meet pressing practical needs. \label{diffi_2} 
    \item There is still a lack of global and convenient (threshold-free) evaluation metrics of fairness for unsupervised anomaly detection methods.  \label{diffi_3}
    \item There is a lack of research on the fairness issues under different data splitting strategies including balanced and skewed data splitting across demographic groups.\label{diffi_4}
\end{enumerate}

In this work, we attempt to address the four difficulties. We present a novel fairness-aware unsupervised AD method called FairAD. FairAD learns to map data from different demographic groups to a common target distribution, which ensures statistical parity for different groups in the target distribution space. The chosen target distribution is supposed to be simple and compact, where simplicity ensures that sampling from the target distribution is easy, and compactness aims to obtain a reliable decision boundary to distinguish between normal and abnormal samples. Furthermore, in order to effectively evaluate fairness in anomaly detection, we propose a novel threshold-free evaluation metric. 
Our main contributions are as follows.
\begin{itemize}
    \item We discuss the group fairness issue and introduce two fundamental assumptions for any fairness-aware (group fairness) UAD methods. Furthermore, we empirically demonstrate that the assumptions are reasonable in real-world scenarios. (for Difficulty~\ref{diffi_1})
    \item We propose FairAD without introducing additional fairness regularization, which achieves a coordinated optimization process for detection task and group fairness and improves the fairness-utility trade-off, in comparison to existing methods. (for Difficulty \ref{diffi_2})
    \item We introduce a threshold-free fairness evaluation metric that holistically quantifies model fairness across the entire decision spectrum, eliminating dependence on manual threshold selection. (for Difficulty \ref{diffi_3})
    \item We consider both balanced and skewed data-splitting strategies across different demographic groups and evaluate all baselines in the two settings. (for Difficulty \ref{diffi_4}) 
\end{itemize}
The experiments on real-world datasets show that our method achieves an improved trade-off between detection accuracy and fairness and the results also verify the effectiveness of the proposed evaluation metrics of fairness. The source code is provided in supplementary materials.

\section{Related Work}

\subsection{Fair Representation Learning}

Fair representation learning (FRL)~\citep{cerrato202410} focuses on mitigating biases and ensuring fairness in machine learning systems by transforming data into latent space where sensitive attributes (e.g., race, gender) have minimal or no influence on outcomes. It aims to achieve equity in predictions across different demographic groups while maintaining task accuracy. The main technical routes include: (1) adversarial learning to promote independence between latent features and sensitive attributes~\citep{xie2017controllable,madras2018learning,zhang2018mitigating}; and (2) mutual information and variational inference~\citep{louizos2015variational,moyer2018invariant,creager2019flexibly,oh2022learning}; and (3) introducing fairness constraints~\cite{agarwal2018reductions}. To some extent, these FRL techniques constitute a key approach to building fair machine learning systems.

Building on the FRL techniques, a straightforward strategy for achieving fair anomaly detection is to construct two-stage pipelines: (1) generating fair embeddings using existing FRL methods, followed by (2) training an anomaly detector on these embeddings. However, this paradigm exhibits two critical limitations in unsupervised anomaly detection scenario: (1) (\textbf{Task Compatibility}) most FRL methods are not disentangled with downstream tasks (typically classification), making them incompatible with unsupervised anomaly detection where all training samples belong to a single (normal) class and lack auxiliary label information, and (2) (\textbf{Detection Efficacy}) on the other hand, the task-agnostic representation is not guaranteed to be useful in distinguishing between normal samples and anomalies, leading to low detection accuracy.  In contrast, our proposed method introduces an end-to-end framework with fairness mechanisms specifically designed for unsupervised anomaly detection. Empirical results (see Section~\ref{Section-Exp}) also demonstrated that our proposed method achieves significantly better detection accuracy than the two-stage pipelines while maintaining comparable fairness.

\subsection{Fairness in Anomaly Detection}
Despite so many works on anomaly detection problems~\citep{Ruff_Vandermeulen_Goernitz_Deecke_Siddiqui_Binder_Müller_Kloft_2018,cai2022perturbation,han2022adbench,bouman2024unsupervised}, the studies on the fairness of anomaly detection are limited. To the best of our knowledge, ~\cite{davidson2020framework} first studied the fairness issue of outlier detection and proposed a framework to determine whether the output of an outlier detection algorithm is fair. Subsequently, \cite{deepak2020fair} studied the fairness problem of LOF (Local Outlier Factor)~\citep{Breunig_Kriegel_Ng_Sander_2000} and proposed a strategy to mitigate the unfairness of LOF on tabular datasets. \cite{zhang2021towards} studied the fairness problem of Deep SVDD~\citep{Ruff_Vandermeulen_Goernitz_Deecke_Siddiqui_Binder_Müller_Kloft_2018} and proposed Deep Fair SVDD, which used an adversarial network to de-correlate the relationships between the sensitive attributes and the learned representations.
\cite{Shekhar_Shah_Akoglu_2021} added statistical parity regularization and group fidelity regularization on AutoEncoder (AE)~\citep{hinton2006reducing} to mitigate the unfairness of AE-based anomaly detection methods. More recently, \cite{han2023achieving} studied counterfactual fairness, which is to ensure the consistency of the detection outcome in the factual and counterfactual world to different demographic groups.

Adversarial training~\citep{zhang2021towards} is known to be unstable, and fairness regularization terms~\citep{Shekhar_Shah_Akoglu_2021} often compromise detection performance, and counterfactual-based methods~\citep{han2023achieving} introduce extra training complexity. In contrast to these methods, our proposed method achieves a simple and coordinated optimization process for the detection task and group fairness. Empirical results (See Section~\ref{Section-Exp}) also demonstrated that our method achieves superior detection accuracy compared to these methods, while maintaining comparable or even better fairness.

\section{Fairness-aware Anomaly Detection}

\subsection{Preliminary Knowledge}\label{sec-3.1}
\paragraph{Unsupervised AD}
Let $\mathcal{X} = \{\mathbf{x}_1, \mathbf{x}_2, \cdots, \mathbf{x}_n\} \subseteq \mathbb{R}^d$ be a set of $n$ samples drawn from an unknown distribution $\mathcal{D}_{\mathbf{x}}$ and the samples from $\mathcal{D}_{\textbf{x}}$ are deemed as normal data. A point $\mathbf{x} \in \mathbb{R}^d$ is deemed to be anomalous if $\mathbf{x}$ is not drawn from $\mathcal{D}_{\textbf{x}}$.  Then, the goal of the unsupervised AD is to obtain a decision function $f: \mathbb{R}^d \rightarrow \{0, 1\}$ by utilizing only $\mathcal{X}$, such that $f(\mathbf{x}) = 0$ if $\mathbf{x}$ is drawn from $\mathcal{D}_{\mathbf{x}}$ and $f(\mathbf{x}) = 1$ if $\mathbf{x}$ is not drawn from $ \mathcal{D}_{\mathbf{x}}$. Note that this is a standard setting of anomaly detection, followed by most unsupervised AD methods~\citep{Ruff_Vandermeulen_Goernitz_Deecke_Siddiqui_Binder_Müller_Kloft_2018, goyal2020drocc, han2022adbench, fu2024dense, xiao2025unsupervised}, where models are trained exclusively on normal data. The main difference among unsupervised AD methods is the design of the decision function $f$. 

\paragraph{Group Fairness}
Demographic parity, a.k.a. statistical parity~\citep{dwork2012fairness}, demands the existence of parity between different demographic groups, such as those defined by gender or race. We use $S\in\mathcal{S}:=\{s_1,s_2,\ldots,s_K\}$ to denote the sensitive or protected attribute and $|\mathcal{S}|=K\geq 2$. There are the following definitions of group fairness.
\begin{definition}[Demographic parity~\citep{agarwal2018reductions}]\label{def_DP_mc}
A predictor $f: \mathcal{X} \rightarrow  \mathcal{Y}$ achieves demographic parity under a distribution over ($\mathbf{x}, S, {y}$) where $y \in \{0, 1\}$ be the data label if its prediction $\hat{y}:=f(\mathbf{x})$ is statistically independent of the protected attribute $S$--- that is, if $\mathbb{P}[f(\mathbf{x}) = \hat{{y}}~|~S=s] = \mathbb{P}[f(\mathbf{x}) = \hat{{y}}]$ for all $s$ and $\hat{{y}}$.
\end{definition}

\begin{definition}[Equal opportunity~\citep{Gajane_2017}] \label{def_EO_mc}
Use the same notations of Definition \ref{def_DP_mc}. A predictor $f$ achieves equal opportunity under a distribution over ($\mathbf{x}, {S}, y$) if its prediction $\hat{y}:=f(\mathbf{x})$ is conditionally independent of the protected attribute ${S}$ given the label $y=1$--- that is, if $\mathbb{P}[\hat{y}=1~|~S=s, y=1] = \mathbb{P}[\hat{y}=1~|~y=1]$ for all $s$.
\end{definition}

In unsupervised AD, $y=0$ and $y=1$ represent normality and anomaly respectively. However, the training data do not include any labeled anomalous samples, which means the predictor $f$ will never be guaranteed to learn sufficient information about the anomaly pattern. In Definition \ref{def_EO_mc} (Equal opportunity), $y=1$ is presented explicitly and in Definition \ref{def_DP_mc} (Demographic parity), $y=1$ is presented implicitly because $\mathbb{P}\left[ f(\mathbf{x}) = \hat{y}~|~S=s\right]$ and $\mathbb{P}\left[ f(\mathbf{x}) = \hat{y}\right]$ depend on both normal samples ($y=0$) and anomalous samples ($y=1$), that is $\mathbb{P}\left[ f(\mathbf{x}) = \hat{y}\right]=\mathbb{P}\left[ f(\mathbf{x}) = \hat{y}~|~y=0\right]\times \mathbb{P}\left[y=0\right]+\mathbb{P}\left[ f(\mathbf{x}) = \hat{y}~|~y=1\right]\times \mathbb{P}\left[y=1\right]$. Therefore, we have the following claim (proved in Appendix~\ref{proof-claim1}).
\begin{claim}
  In unsupervised AD, neither demographic parity nor equal opportunity can be meaningfully guaranteed if without any additional assumptions.  
\end{claim}

Note that ``meaningfully'' emphasizes the trivial solutions $f(\mathbf{x})\equiv 0~\text{or}~1$ for any $\mathbf{x}$ are excluded. In fact, without additional assumptions, any fairness involving anomalous data cannot be guaranteed in unsupervised settings.
It is worth noting that \cite{Shekhar_Shah_Akoglu_2021} considered the equal opportunity in unsupervised AD, where a fairness-unaware AD model (base model) is utilized first to predict pseudo-label $\hat{y}$ and then they used $\hat{y}$ to ensure equal opportunity. Obviously, such a strategy has a significant limitation: the result of equal opportunity depends on the detection performance of the base model, of which fairness cannot be guaranteed. Therefore, for unsupervised AD, without additional assumptions, it is only possible to guarantee the following fairness.
\begin{definition}[Predictive equality \citep{chouldechova2017fair}] Let $\hat{y}:=f(\mathbf{x})$. The false positive error rate balance, a.k.a. predictive equality, is defined as
\begin{equation}\label{eq_PE}
\mathbb{P}\left[ \hat{y}=1~|~S=s_i,y=0\right] =\mathbb{P}\left[\hat{y}=1~|~S=s_j,y=0\right]
\end{equation}
where $s_i,s_j\in \mathcal{S},~ i\neq j$.
\end{definition}

\subsection{Fairness Discussion in Unsupervised Anomaly Detection}

As evidenced by the preceding analysis, it is intractable to ensure group fairness both on normal and abnormal data in unsupervised anomaly detection (UAD).
However, existing works~\citep{zhang2021towards,Shekhar_Shah_Akoglu_2021,han2023achieving} have empirically demonstrated that group fairness (demographic parity or equal opportunity) can be achieved to some extents by introducing adversarial training, adding fairness constraints to optimization objective or generating counterfactual samples for training. This suggests that there must be some natural conditions implicitly contributing to group fairness in unsupervised anomaly detection.
In real-world scenarios, anomalous instances are not completely unrelated to normal instances; otherwise, the detection task would be trivial. It is quite common that anomalous samples emerge as perturbed normal samples and the evolution from normality to anomaly is gradual. For instance, in chemical engineering, flow control valves will gradually block, leading to failure; in the mechanical field, bearings will gradually deform, resulting in abnormal vibration signals. Meanwhile, some normal samples are naturally close to anomalous samples. Therefore, it is possible to learn some patterns of anomaly from the normal training data and we make the following assumption. 

\begin{assumption}[Learnable abnormality]\label{assump_1}
Suppose $\mathcal{X}=\{\mathbf{x}_1, \mathbf{x}_2, \cdots, \mathbf{x}_n\}$ are independently drawn from $\mathcal{D}_{\mathbf{x}}$.
The abnormalities of both $\mathcal{X}$ and unknown anomalous samples can be correctly quantified by a function $\mathcal{T}^*:\mathbb{R}^d \rightarrow \mathbb{R}$, and there exists a permutation $\pi$ on $\mathcal{X}$ such that $0 \leq \mathcal{T}^*(\mathbf{x}_{\pi_1})<\mathcal{T}^*(\mathbf{x}_{\pi_2})<\cdots<\mathcal{T}^*(\mathbf{x}_{\pi_n})$, where a larger value of $\mathcal{T}^*(\mathbf{x})$ means that $\mathbf{x}$ is not drawn from the normal distribution $\mathcal{D}_{\mathbf{x}}$ with a higher probability.
\end{assumption}

Assumption~\ref{assump_1} reflects the fundamental expectation that normal data should be compact in the feature space. Formally, this requires the existence of a bounding function $\mathcal{T}^*: \mathbb{R}^d \rightarrow \mathbb{R}$ such that observed normal samples satisfy $\mathcal{T}^*(\mathbf{x}_{\text{normal}}) < \mathcal{T}^*(\mathbf{x}_{\text{abnormal}})$. This assumption ensures the learnability of normal patterns while providing feasibility guarantees for the anomaly detection task.
Based on Assumption~\ref{assump_1}, $\mathcal{T}^*$ servers as an anomaly score function and performs extrapolation on anomalous samples. Most existing UAD methods~\citep{han2022adbench,bouman2024unsupervised} have demonstrated that it is possible to learn an approximation of $\mathcal{T}^*$ from $\mathcal{X}$ that can generalize to unseen anomalous samples. However, $\mathcal{T}^*(\mathbf{x})$ may not be independent from a sensitive attribute $S$ and hence can lead to unfairness. 

To achieve fairness on $\mathcal{X}$, we split $\mathcal{X}$ into different protected groups $\mathcal{X}_{S=s_i}$ according to the values of a sensitive attribute $S$. Therefore, a fair AD model $\hat{y}=f(\mathbf{x})$ on $\mathcal{X}$ is to ensure
\begin{equation}\label{eq_yxh}
\begin{aligned}
     & \mathbb{P}[\hat{y} ~|~\mathbf{x} \in \mathcal{X}_{S=s_i}, \mathcal{T}^*(\mathbf{x}_{\pi_1})\leq \tilde{y} \leq \mathcal{T}^*(\mathbf{x}_{\pi_n})] \\
     =~& \mathbb{P}[\hat{y}~|~\mathbf{x} \in \mathcal{X}_{S=s_j},\mathcal{T}^*(\mathbf{x}_{\pi_1})\leq \tilde{y}\leq \mathcal{T}^*(\mathbf{x}_{\pi_n})]
\end{aligned}
\end{equation}
where $\tilde{y}:=\mathcal{T}^*(\mathbf{x})$ denotes the anomaly level of $\mathbf{x}$.
Different from \eqref{eq_PE} that does not involve any information about the anomaly, \eqref{eq_yxh} is associated with $\mathcal{T}^*$ (established in Assumption~\ref{assump_1}), meaning that it is possible to learn some information about the anomaly from $\mathcal{X}$. Therefore, it is reasonable to make the following assumptions for fairness on abnormal samples, where $E:=\big\vert \mathbb{P}[\hat{y} ~|~\mathbf{x} \in \mathcal{X}_{S=s_i}, \mathcal{T}^*(\mathbf{x}_{\pi_1})\leq \tilde{y} \leq \mathcal{T}^*(\mathbf{x}_{\pi_n})]-\mathbb{P}[\hat{y}~|~\mathbf{x} \in \mathcal{X}_{S=s_j},\mathcal{T}^*(\mathbf{x}_{\pi_1})\leq \tilde{y}\leq \mathcal{T}^*(\mathbf{x}_{\pi_n})]\big\vert$.

\begin{assumption}[Transferable fairness]\label{assump_2}
    Let $\tilde{\mathcal{X}}$ be the set of all unseen anomalous samples and $\tilde{E}=\big\vert \mathbb{P}[\hat{y} ~|~\mathbf{x} \in \tilde{\mathcal{X}}_{S=s_i}, ~\tilde{y}>\mathcal{T}^*(\mathbf{x}_{\pi_n})]-\mathbb{P}[\hat{y}~|~\mathbf{x} \in \tilde{\mathcal{X}}_{S=s_j},~\tilde{y}>\mathcal{T}^*(\mathbf{x}_{\pi_n})]\big\vert$. There exists a small constant $\mu\geq 1$ such that
    $\tilde{E}\leq \mu E$.
\end{assumption}
\begin{assumption}[Generalizable parity]\label{assump_3}
    Let $\mathcal{\hat{X}}$ be the union of the set of all unseen anomalous samples and the set of all unseen normal samples and $\hat{E}=\big\vert \mathbb{P}[\hat{y} ~|~\mathbf{x} \in \mathcal{\hat{X}}_{S=s_i}, 0\leq \tilde{y}]-\mathbb{P}[\hat{y}~|~\mathbf{x} \in \mathcal{\hat{X}}_{S=s_j},0\leq \tilde{y}]\big\vert$. There exists a small constant $\tau\geq 1$ such that
    $\hat{E}\leq \tau E$.
\end{assumption}
Assumption~\ref{assump_2} ensures that when $\mathcal{T}^*$ is fair on the training data, it is also fair on unseen anomalous data, provided that $\mu$ is not too large, where $\mu$ depends on real data distribution and the unknown score function $\mathcal{T}^*$. If $\tilde{E}=0$ for any $i,j$, the assumption implies equal opportunity. Similarly, Assumption~\ref{assump_3} implies demographic parity when $\hat{E}=0$. 
Therefore, the evaluation of a fair AD method depends on both the fairness principle (e.g., demographic parity or equal opportunity) and assumption. 
Indeed, Assumptions~\ref{assump_2},~\ref{assump_3} have already implicitly verified by existing fairness-aware AD methods~\citep{zhang2021towards,Shekhar_Shah_Akoglu_2021,han2023achieving}. Otherwise, it is only possible to guarantee the predictive equality~\eqref{eq_PE}. In our experiments, the results on real-world datasets validate the reasonability of the two assumptions again.

\subsection{Model Formulation}

\subsubsection{Anomaly Detection via Compact Distribution Transformation}
\label{UAD-method}
For unsupervised anomaly detection, density estimation~\citep{silverman2018density} is an effective strategy~\citep{zong2018deep, Ruff_Vandermeulen_Goernitz_Deecke_Siddiqui_Binder_Müller_Kloft_2018,fu2024dense} to distinguish between normal and abnormal instances. However, the dimensionality of the data is often high and the data distribution in the original space is complex, which makes density estimation challenging. 

To solve this problem, we propose to learn a projection $\mathcal{P}: \mathbb{R}^d \rightarrow \mathbb{R}^m$ that transforms data distribution $\mathcal{D}_{\textbf{x}}$ to a known target distribution $\mathcal{D}_{\textbf{z}}$ that is simple and compact, while there still exists a projection $\mathcal{P}^\prime: \mathbb{R}^m \rightarrow \mathbb{R}^d$ that can recover $\mathcal{D}_\mathbf{x}$ from $\mathcal{D}_\mathbf{z}$ approximately, ensuring that the major information of $\mathbf{x}$ can be preserved by $\mathbf{z}$, where $\mathbf{z}\sim\mathcal{D}_{\mathbf{z}}$. Therefore, we aim to solve
\begin{equation}
\begin{aligned}
    \underset{\mathcal{P}, \mathcal{P}^\prime}{\text{min}}~& \mathcal{M}(\mathcal{P}(\mathcal{D}_\mathbf{x}), \mathcal{D}_{\mathbf{z}}) \\
    \text{s.t.} &~\mathcal{M}(\mathcal{P}^\prime(\mathcal{P}(\mathcal{D}_\mathbf{x})), \mathcal{D}_{\mathbf{x}})\leq c
\end{aligned}
    \label{eq3-mc}
\end{equation}
where $\mathcal{M}(\cdot, \cdot)$ denotes a distance metric between distributions and $c$ is some positive constant. Instead of the constrained optimization problem \eqref{eq3-mc}, we can simply solve the following problem
\begin{equation}
    \underset{\mathcal{P}, \mathcal{P}^\prime}{\text{min}} ~\mathcal{M}(\mathcal{P}(\mathcal{D}_\mathbf{x}), \mathcal{D}_{\mathbf{z}}) + 
    \beta \mathcal{M}(\mathcal{P}^\prime(\mathcal{P}(\mathcal{D}_\mathbf{x})), \mathcal{D}_{\mathbf{x}}),
    \label{eq3.5}
\end{equation}
where $\beta>0$ is a trade-off hyperparameter for the two terms. 
We use two deep neural networks $h_\phi$ and $g_\psi$ with parameters $\phi, \psi$ to model $\mathcal{P}$ and $\mathcal{P}^\prime$ respectively. Now, the problem~\eqref{eq3.5} becomes
\begin{equation}
    \underset{\phi, \psi}{\text{min}} ~ \mathcal{M}(\mathcal{D}_{h_\phi(\mathbf{x})}, \mathcal{D}_{\mathbf{z}}) + 
    \beta \mathcal{M}(\mathcal{D}_{g_\psi(h_\phi(\mathbf{x}))}, \mathcal{D}_{\mathbf{x}})
    \label{eq4-mc}
\end{equation} 
However, problem~\eqref{eq4-mc} is intractable as data distribution $\mathcal{D}_\mathbf{x}$ is unknown and $\mathcal{D}_{h_\phi(\mathbf{x})}, \mathcal{D}_{g_\psi(h_\phi(\mathbf{x}))}$ cannot be computed analytically. Thus, we expect to measure the distance between distributions using their finite samples because we can sample from $\mathcal{D}_\mathbf{x}$ and $\mathcal{D}_\mathbf{z}$ easily. A feasible and popular choice of $\mathcal{M}(\cdot, \cdot)$ is the Sinkhorn distance~\citep{cuturi2013sinkhorn} between two distributions supported by $\mathcal{X}=\{\mathbf{x}_1,\mathbf{x}_2, \ldots, \mathbf{x}_{n_1}\}$ and $\mathcal{Y} = \{\mathbf{y}_1,\mathbf{y}_2, \ldots, \mathbf{y}_{n_2}\}$:
\begin{equation}
\begin{aligned}
     \text{Sinkhorn}(\mathcal{X}, \mathcal{Y}):=  \min& ~\langle \mathbf{P}, \mathbf{C} \rangle_F + \alpha \sum_{i,j}\mathbf{P}_{ij}\log(\mathbf{P}_{ij}), \\
     & \text{s.t.}~\mathbf{P}\mathbf{1} = \mathbf{a}, \mathbf{P}^T\mathbf{1}=\mathbf{b}, \mathbf{P} \geq 0
     \label{eq-sinkhorn}
\end{aligned}
\end{equation}
where $\mathbf{P}\in\mathbb{R}^{n_1\times n_2}$ denotes the transport plan and $\mathbf{C}\in\mathbb{R}^{n_1\times n_2}$ is a metric cost matrix between $\mathcal{X}$ and $\mathcal{Y}$. The two probability vectors $\mathbf{a}$ and $\mathbf{b}$ satisfy $\mathbf{a}^T\mathbf{1}=1, \mathbf{b}^T\mathbf{1}=1, \mathbf{1}=[1,1,\cdots, 1]^T$. Other measures, such as the maximum mean discrepancy, can also be used, which is discussed in the Appendix~\ref{com-sinkhorn-mmd}.   
Now, we use Sinkhorn distance to replace the first term in problem~\eqref{eq4-mc}, use reconstruction error to replace the second term in problem~\eqref{eq4-mc}, and get the following optimization objective
\begin{equation}
    \underset{\phi, \psi}{\text{min}} ~ \text{Sinkhorn}(h_\phi(\mathcal{X}), \mathcal{Z}) + \frac{\beta}{n}\sum_{i=1}^{n} \Vert \mathbf{x}_i - g_\psi(h_\phi(\mathbf{x}_i)) \Vert^2
    \label{eq5}
\end{equation}
where $\mathbf{x}_i \in \mathcal{X}$, $\mathcal{Z}=\{\mathbf{z}_i : \mathbf{z}_i \sim \mathcal{D}_z, i=1, \dots, {n}\}$, and $\beta$ is a trade-off hyperparameter. 

We may replace both the first and second terms in problem~\eqref{eq4-mc} using Sinkhorn distance, however, which easily leads to a higher computational cost. Therefore, we use reconstruction error to replace the second term in problem~\eqref{eq4-mc}. The reconstruction error term is a stronger constraint than $\mathcal{P}^\prime$, but it is efficient and effective to preserve the core information from the original data distribution $\mathcal{D}_\mathbf{x}$.  

\paragraph{Target distribution} The target distribution should be simple and compact. The compactness ensures that projected normal samples in the decision space lie in high-density regions, which contributes to a reliable decision boundary. The simplicity ensures that sampling from the target distribution is easy.
Therefore, for the target distribution $\mathcal{D}_{\mathbf{z}}$, we propose to use a \textit{truncated isotropic Gaussian} based on $\mathcal{N}(\mathbf{0},\mathbf{I}_d)$ where truncation is to ensure that the target distribution is sufficiently compact for normal data. Based on the target distribution, we can define a score function naturally depending on density estimation.

\paragraph{Anomaly score} Let $h_{\phi^*}$ be the trained fairness-aware model. In inference phase, for a new sample $\mathbf{x}_{\text{new}}$, we define a soft anomaly score $\zeta_{\text{soft}}$ (or a hard score depending on a threshold)
\begin{equation} \label{def_score}
      \text{Score}(\mathbf{x}_{\text{new}}) = \zeta_{\text{soft}}(h_{\phi^\ast}(\mathbf{x}_{\text{new}})) = \Vert h_{\phi^\ast}(\mathbf{x}_{\text{new}})\Vert,
\end{equation}
% \begin{equation}
%     \text{Score}(\mathbf{x}_{\text{new}}) = \Vert h^*_\phi(\mathbf{x}_{\text{new}})\Vert,
% \end{equation}
which can measure the anomaly degree of $\mathbf{x}_{\text{new}}$. By considering a threshold (with a certain significance level) obtained from the training data scores, we get a hard score function $\zeta_{\text{hard}}$ with a binary output in $\{0,1\}$, indicating whether $\mathbf{x}_{\text{new}}$ is normal or not. Actually, the density of $\mathbf{x}_{\text{new}}$ can be estimated as $(2 \pi)^{-d/2}\exp\left(-\frac{1}{2}(\text{Score}(\mathbf{x}_{\text{new}}))^2\right)$.
% Actually, the density of $\mathbf{x}_{\text{new}}$ can be estimated as
% $(2 \pi)^{-d/2}\exp\left(-\frac{1}{2}\text{Score}^2(\mathbf{x}_{\text{new}})\right)$.

\subsubsection{Fairness via Shared Target Distribution}
Directly finding a fair~\footnote{Unless specified otherwise, all ``fair" and ``fairness" in this paper pertain to group fairness.} $f$ while maintaining strong detection ability is non-trivial. A naive strategy involves combining FRL techniques with UAD methods to construct two-stage pipelines, in which the optimization objectives in the FRL stage are not tailored for the detection task. As a result, Such pipelines often yield poor detection performance. Similarly, directly incorporating fairness constraints into the optimization objectives of UAD methods also degrades detection performance. To avoid such problems, we expect to find an end-to-end and coordinated learning process for the detection task and group fairness. Based on the framework established in Section~\ref{UAD-method}, we can obtain a projection $h_{\phi^*}$ via compact distributional transformation, and then the detection task can be conducted effectively by $\zeta$ estimating the density of data in decision space. Thus, we obtain a detector $f =\zeta\circ h$ and $\hat{y} = f(\mathbf{x})=\zeta\circ h(\mathbf{x})$.
Notably, if $h$ is fair for a protected variable $S$, we have
\begin{equation}\label{eq_zxh}
\begin{aligned}
     & \mathbb{P}[h(\mathbf{x}) ~|~\mathbf{x} \in \mathcal{X}_{S=s_i}, \mathcal{T}^*(\mathbf{x}_{\pi_1})\leq \tilde{y}\leq \mathcal{T}^*(\mathbf{x}_{\pi_n})] \\
     =~& \mathbb{P}[h(\mathbf{x})~|~\mathbf{x} \in \mathcal{X}_{S=s_j},\mathcal{T}^*(\mathbf{x}_{\pi_1})\leq \tilde{y}\leq \mathcal{T}^*(\mathbf{x}_{\pi_n})].    
\end{aligned}
\end{equation}
where $s_i, s_i \in S$ denote the attribute values of protected variable $S$.

It is easy to show that (proved in Appendix~\ref{proof-prop1})
\begin{proposition}\label{prop_zeta_equiv}
    For any $\zeta$, if \eqref{eq_zxh} is attained, then \eqref{eq_yxh} holds.
\end{proposition}

To obtain a fair $h$ without introducing additional fairness constraints, we propose to map data across different demographic groups into a common target distribution $\mathcal{D}_\mathbf{z}$. Naturally, the \eqref{eq3-mc} becomes

\vspace{-10pt}
\begin{equation}
\begin{aligned}
    \underset{\mathcal{P}, \mathcal{P}^\prime}{\text{min}}~& \sum_{s \in S} \mathcal{M}(\mathcal{P}(\mathcal{D}_{\mathcal{X}_{S=s}}), \mathcal{D}_{\mathbf{z}}) \\
    \text{s.t.} &~\mathcal{M}(\mathcal{P}^\prime(\mathcal{P}(\mathcal{D}_\mathbf{x})), \mathcal{D}_{\mathbf{x}})\leq c
\end{aligned}
    \label{eq3-f-mc}
\end{equation}

It further follows that (proved in Appendix~\ref{proof-prop2})
\begin{proposition}\label{prop_equiv_2}
    If $\sum_{s \in S} \mathcal{M}(\mathcal{P}(\mathcal{D}_{\mathcal{X}_{S=s}}), \mathcal{D}_{\mathbf{z}})=0$, then \eqref{eq_zxh} is attained.
\end{proposition}

\begin{wrapfigure}{r}{0.5\textwidth}
\vspace{-15pt}
    \centering
    \includegraphics[width=1\linewidth]{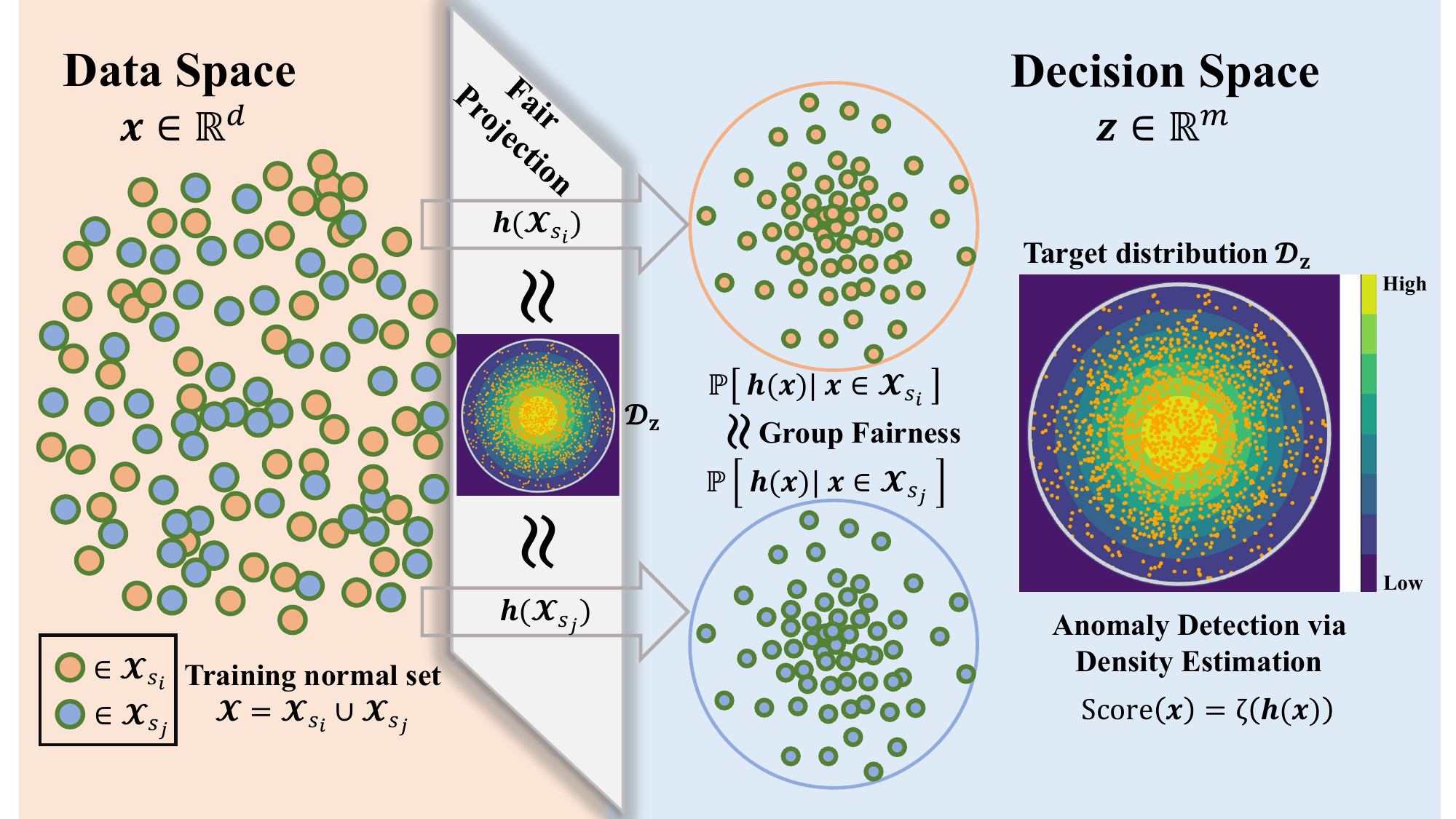}
    \caption{The illustration of Im-FairAD. For simplicity, we only visualize two different attribute values $s_i, s_j \in S$ for the protected variable $S$, but Im-FairAD does not impose such a restriction. The `High' and `Low' denote the relative density in the target distribution.}
    \label{fig-fairness}
    \vspace{-15pt}
\end{wrapfigure}

Therefore, combing Proposition \ref{prop_zeta_equiv} and Proposition \ref{prop_equiv_2}, we conclude that solving problem~\eqref{eq3-f-mc} makes the decision function $f$ as fair as possible on the training data, in terms of predictive equality. Further, using Assumption \ref{assump_2} or \ref{assump_3}, we may obtain equal opportunity or demographic parity. 

Based on the above analysis, we obtain a projection $h$ via compact distribution transformation (common target distribution for different demographic groups), where $h$ can satisfy different fairness principles building on different assumptions, and meanwhile yield a high-detection-accuracy detector $f=\zeta \circ h$ based on a compact target distribution. Building on the framework in Section~\ref{UAD-method}, we finally solve 

\vspace{-10pt}

\begin{equation}
    \underset{\phi, \psi}{\text{min}} ~ \sum_{s \in S}\text{Sinkhorn}(h_\phi(\mathcal{X}), \mathcal{Z}) + \frac{\beta}{n}\sum_{i=1}^{n} \Vert \mathbf{x}_i - g_\psi(h_\phi(\mathbf{x}_i)) \Vert^2.
    \label{eq5-f-mc}
\end{equation}

Notably, our approach achieves fairness across demographic groups of protected variables without introducing additional fairness constraints, thereby fulfilling the dual objectives of detection efficacy and group fairness. We call the method \textbf{Im-FairAD}. Figure~\ref{fig-fairness} provides an illustration. 

\vspace{-5pt}
\paragraph{Variant of FairAD} Besides the method presented by~\eqref{eq5-f-mc}, we also explore another feasible optimization objective by directly minimizing the distribution distance of anomaly scores across different groups, i.e.,
\vspace{-10pt}
\begin{equation}
\begin{aligned}
        \underset{\phi, \psi}{\text{min}} ~~& \text{Sinkhorn}(h_\phi(\mathcal{X}), \mathcal{Z}) + \frac{\beta}{n}\sum_{i=1}^n \Vert \mathbf{x}_i - g_\psi(h_\phi(\mathbf{x}_i)) \Vert^2 \\
        & + \lambda\sum_{i\neq j}\text{Sinkhorn}(\zeta(h_\phi(\mathcal{X}_{S=s_i})),\zeta(h_\phi(\mathcal{X}_{S=s_j}))).
\end{aligned}
    \label{eq6}
\end{equation}
The first and second terms focus on detection accuracy and the third term ensures fairness between protected groups. We call the method \textbf{Ex-FairAD}. It is worth noting that the first term in~\eqref{eq6}, mapping data distribution into target distribution, is necessary for Ex-FairAD because it determines whether the score function~\eqref{def_score} is feasible for Ex-FairAD. More detailed discussion and an ablation study are provided in Appendix~\ref{abalation-study}.

\subsection{Threshold-Free Fairness Metrics}
 
To overcome the limitations (See Appendix~\ref{dis-adpd-fr}) of \textit{fairness ratio}, in this paper, we propose a new fairness metric called \textit{Average Demographic Parity Difference} (ADPD):
\begin{equation}
    \text{ADPD} := \frac{1}{n}\sum_{k=1}^{n} \Big\vert  \mathbb{P}(\text{Score}(\mathcal{X}) > t_k | S=s_i ) 
    -  \mathbb{P}(\text{Score}(\mathcal{X}) > t_k | S=s_j) \Big\vert
    \label{eq7-1}
\end{equation}
where $t_k \in \text{Score}(\mathcal{X})$ denotes the anomaly score of single sample. In our proposed methods, $t_k = \Vert h_{\phi^\ast}(\mathbf{x}_k)\Vert$. ADPD is a threshold-free metric (such as AUC) for measuring demographic parity. The range of $\text{ADPD}$ is $[0, 1)$ and a smaller ADPD means a higher fairness. Although we introduce a novel threshold-free metric for fairness measure, this is not to imply that the threshold-dependent metrics are useless. More discussion on threshold-free and threshold-dependent metrics is provided in Appendix~\ref{dis-adpd-fr}. We both use threshold-free and threshold-dependent metrics to evaluate all methods in our experiments.  

\section{Experiments}
\label{Section-Exp}

Our experiments are conducted on six publicly available datasets from the literature on fairness in machine learning ~\citep{zhang2021towards,chai2022self,han2023achieving,chen2024fairness}, where there are different kinds of sensitive information. The more detailed statistics of datasets and data splitting are provided in Appendix~\ref{data-baselines}. We both use standard (fairness-unaware) UAD methods, two-stage pipelines (FRL technique + UAD methods), and end-to-end fairness-aware UAD methods as baselines. 
% Note that for all experiments, only normal samples are used in the training stage, but there are both normal and abnormal samples in the test stage. 
The detailed experimental settings are provided in Appendix~\ref{exp-settings}. We design experiments to answer the following questions.

\begin{itemize}
    \item \textbf{[Q1]} Can our proposed methods achieve better detection accuracy when maintaining comparable or even better fairness?
    \item \textbf{[Q2]} How are the performances (including detection accuracy and fairness) of all compared methods on balanced and skewed splitting?
    \item \textbf{[Q3]} Can fairness-aware unsupervised AD methods achieve fairness on anomalous data in real-world scenarios, although anomalous samples are not used during the training?
    
\end{itemize}

\subsection{Experimental Results}

\begin{figure*}[h!]
    \vspace{-8pt}
    \centering
    \includegraphics[width=\textwidth]{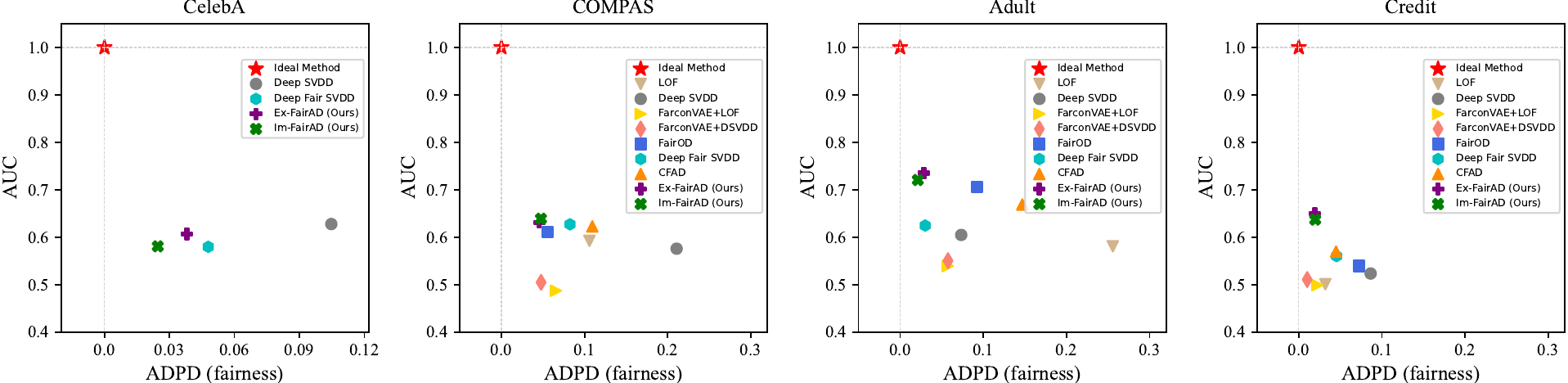}
    \caption{Accuracy-fairness trade-off on balanced splitting. Note that the baselines, FairOD and CFAD are tailored to tabular data.}
    \label{fig-auc-vs-adpd}
    \vspace{-10pt}
\end{figure*}

To answer \textbf{[Q1]}, we visualize the trade-off between AUC and ADPD under balanced splitting (same sample size across different demographic groups) in Figure~\ref{fig-auc-vs-adpd}, where the red star in the upper left corner denotes the ideal fairness-aware anomaly detection. Compared with all the baselines, our method achieves comparable or even better trade-off between detection accuracy and fairness in almost all cases. The ADPD in Figure~\ref{fig-auc-vs-adpd} is computed on all test data, including normal and abnormal samples. The detailed numerical results and trade-off visualizations under skewed splitting (varying sample size across different demographic groups) are provided in Appendix~\ref{res-adpd-auc}.

Further, the empirical results in Table~\ref{tab-adpd-auc-cacc} reveal a critical trend: when transitioning from balanced to skewed splitting, fairness-aware AD methods exhibit degradation in group fairness, as evidenced by significantly larger ADPD values in most cases. This suggests that fairness of existing fairness-aware AD methods is easily affected by demographic imbalances and skewed sample proportions across groups amplify fairness violations. By contrast, our proposed methods exhibit a robust fairness preservation (slight fluctuation on ADPD) on balanced and skewed data splitting~\textbf{[Q2]}.

\textbf{Fairness of Im-FairAD under balanced and skewed splitting}
When problem~\eqref{eq-sinkhorn} is solved well that is $\sum_{s \in S} \mathcal{M}(h(\mathcal{D}_{\mathcal{X}_{S=s}}), \mathcal{D}_{\mathbf{z}})$ near to zero, whatever under balanced or skewed data splitting, Im-FairAD can obtain fair $h(\mathcal{X}_{S=s})$ for different groups $s$. Therefore, Im-FairAD can guarantee~\eqref{eq_zxh} according to Proposition~\ref{prop_equiv_2} for both balanced and skewed data splitting \textbf{[Q2]}.

\textbf{Fairness of Im-FairAD on abnormal data}

\begin{wrapfigure}{r}{0.5\textwidth}
% \vspace{-5pt}
    \centering
    \includegraphics[width=1\linewidth]{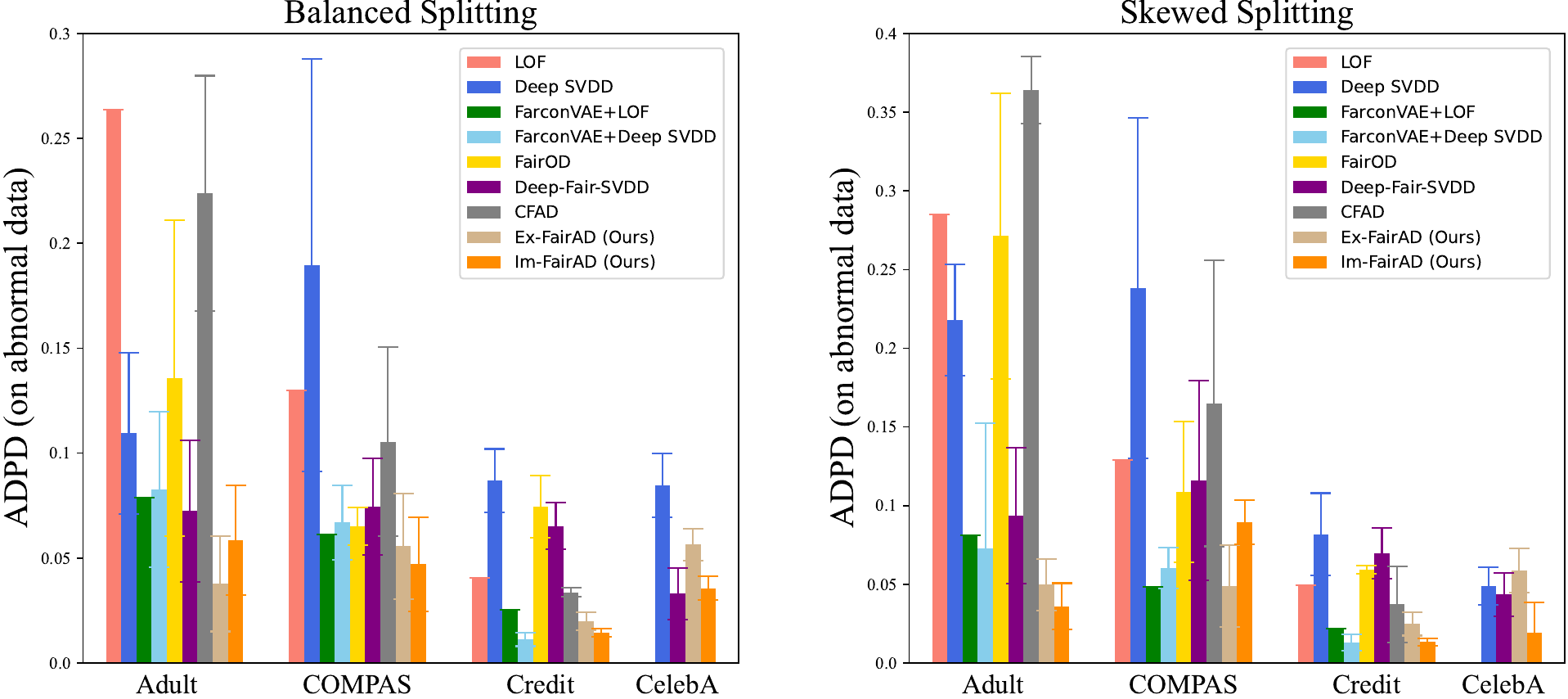}
    \caption{Fairness of all baselines on abnormal data from the test set.}
    \label{fig-abnormal-adpd}
\vspace{-5pt}
\end{wrapfigure}
As there are no abnormal samples in the training set, the optimization processes of unsupervised fairness-aware AD methods do not exploit any information directly from the abnormal data, and hence may not consider the fairness in abnormal data explicitly. 
To answer \textbf{[Q3]}, we visualize the ADPD of abnormal data from the test set in Figure~\ref{fig-abnormal-adpd}. More results on Titanic and SP are provided in Appendix~\ref{res-adpd-auc}. Compared to all baselines, our proposed methods achieve better or comparable group fairness in most cases. On the other hand, Deep Fair SVDD achieves better group fairness than Deep SVDD (fairness-unaware) on abnormal data of all four datasets. Two-stage pipelines (FarconVAE+LOF and FarconVAE+Deep SVDD) also achieve better group fairness than LOF and Deep SVDD (fairness-unaware) on abnormal data. On COMPAS and Credit, FairOD and CFAD both achieve better group fairness than Deep SVDD(fairness-unaware) on abnormal data. One possible reason for the success of these methods especially ours is that the abnormal data may have some similar latent structure as the normal data, or at least there exists a mapping (not too complex)\footnote{This assumption is realistic because in many real scenarios, abnormality origins from normality.} between the normal data distribution $\mathcal{D}_{\mathbf{x}}$ and the abnormal data distribution $\mathcal{D}_{\tilde{\mathbf{x}}}$, i.e., $\mathcal{D}_{\tilde{\mathbf{x}}}=\mathcal{Q}(\mathcal{D}_{\mathbf{x}})$. Thus, the AD methods can ensure fairness on abnormal data indirectly, where the intermediary is the normal data, which also indicates that the \textbf{Assumption}~\ref{assump_2} is reasonable and can be accessed in real-world scenarios. Particularly, our methods are based on distribution transformation and the fairness of normal data can be transformed to abnormal data via the composition $\mathcal{P}\circ\mathcal{Q}$. That is also why our methods are more effective in terms of fairness than other methods on abnormal data.

\subsection{Evaluation by Threshold-Dependent Metric}
\label{measure-fair-ra}
We also use \textit{fairness ratio}~\eqref{eq7} from previous works~\citep{zhang2021towards,Shekhar_Shah_Akoglu_2021} and F1-score to evaluate fairness and detection accuracy. For calculating \textit{fairness ratio} and F1-score, a threshold needs to be determined. We sort the anomaly scores of the training set in ascending order and set the threshold to $pN$-th smallest anomaly score, where we set $p=\{0.90, 0.95\}$ and $N$ denotes the size of the training set. The results on COMPAS, Adult and Credit with $p=0.90$ are reported in Table~\ref{tab-compas-adult-fr-f1}. More results are reported in Appendix~\ref{res-fr-f1}. Observing the Table~\ref{tab-compas-adult-fr-f1}, our methods still achieve better detection accuracy while maintaining a better or comparable \textit{fairness ratio} in almost all cases \textbf{[Q1]}. Notably, the \textit{fairness ratio} is highly sensitive to different thresholds (See all results with $p=\{0.90, 0.95\}$ in Appendix~\ref{res-fr-f1}). Consequently, a threshold-free fairness metric is essential for evaluating the performance from a holistic view.

\begin{table}[h!]
\vspace{-10pt}
    \centering
    \caption{F1-score and \textit{fairness ratio} on COMPAS, Adult and Credit with $p=0.90$. The best two results are marked in \textbf{bold}.}
    \label{tab-compas-adult-fr-f1}
    \resizebox{\columnwidth}{!}{
    \begin{tabular}{l|c|c|c|c|c|c|c|c|c}
    \toprule
    \multirow{3}{*}{Methods} & \multicolumn{3}{c|}{COMPAS} & \multicolumn{3}{c|}{Adult} & \multicolumn{3}{c}{Credit}  \\
    \cline{2-10}
    & \multirow{2}{*}{F1(\%)~$\uparrow$} & \multicolumn{2}{c|}{\textit{fairness ratio}~$\uparrow$} & \multirow{2}{*}{F1(\%)~$\uparrow$} & \multicolumn{2}{c|}{\textit{fairness ratio}~$\uparrow$} &   \multirow{2}{*}{F1(\%)~$\uparrow$} & \multicolumn{2}{c}{\textit{fairness ratio}~$\uparrow$}\\
    \cline{3-4}\cline{6-7}\cline{9-10}
    & & normal & all & & normal & all & & normal & all\\
    \midrule
    LOF & 29.85(0.00) & 0.27(0.00) & 0.35(0.00) & 27.45(0.00) & 0.34(0.00) & 0.42(0.00) & 20.11(0.00) & 0.45(0.00) & 0.55(0.00)\\
    Deep SVDD & 31.30(5.17) & 0.49(0.18) & 0.58(0.18) & 44.73(4.72) & 0.54(0.24) & 0.63(0.11) & 23.31(7.75) & 0.76(0.12) & 0.74(0.20)\\
    FarconVAE+LOF & 14.28(0.00) & 0.69(0.00) & \textbf{0.77}(0.00) & 22.87(0.00) & 0.63(0.00) & 0.62(0.00) & 15.06(0.00) & \textbf{0.88}(0.00) & 0.88(0.00)\\
    FarconVAE+Deep SVDD & 19.50(0.89) & \textbf{0.74}(0.07) & 0.70(0.14) & 22.08(8.87) & 0.59(0.10) & 0.73(0.18) & 16.96(1.80) & \textbf{0.94}(0.02) & \textbf{0.95}(0.02)\\
    FairOD & 17.81(0.11) & 0.70(0.04) & \textbf{0.79}(0.01) & 36.87(3.99) & 0.70(0.06) & 0.32(0.04) & 41.00(0.81) & 0.70(0.04) & 0.74(0.02)\\

    Deep Fair SVDD & 42.23(0.81) & 0.46(0.04) & 0.58(0.04) & 45.03(1.56) & \textbf{0.86}(0.01) & 0.83(0.05) & 13.54(0.75) & 0.81(0.06) & 0.73(0.09)\\

    CFAD & 40.67(3.67) & 0.39(0.13) & 0.49(0.08) & 48.01(5.51) & 0.76(0.08) & 0.55(0.14) & 20.43(0.75) & 0.69(0.05) & 0.69(0.04)\\
    \midrule
    Ex-FairAD (Ours) & \textbf{42.30}(5.88) & \textbf{0.71}(0.18) & 0.74(0.15) & \textbf{52.75}(2.19) & \textbf{0.87}(0.11) & \textbf{0.92}(0.03) & \textbf{53.70}(2.53) & 0.79(0.03) & \textbf{0.90}(0.04)\\
    Im-FairAD (Ours) & \textbf{47.49}(4.32) & 0.65(0.28) & 0.68(0.09) & \textbf{56.04}(4.70) & 0.72(0.06) & \textbf{0.84}(0.07) &\textbf{54.32}(3.43) & 0.81(0.08) & \textbf{0.90}(0.04)\\
    \bottomrule
    \end{tabular}}
\vspace{-10pt}
\end{table}

\subsection{More Experimental Results and Analysis}
Due to space limitations, the appendices contain more results and further experimental investigation.
Appendix~\ref{more-results}: More numerical results and visualization;
Appendix~\ref{abalation-study}: Ablation study on the proposed methods (Im-FairAD and Ex-FairAD);
Appendix~\ref{ana-time-complexity}: Time complexity analysis for proposed method;
Appendix~\ref{eo-effectiveness}: The effectiveness analysis of the proposed methods on equal opportunity;
Appendix~\ref{exp-contaminated}: Experimental investigation on contaminated training set (including unknown abnormal samples).

\section{Conclusion}
In this paper, we focused on the group fairness of unsupervised anomaly detection, clearly discussing the necessary conditions of achieving group fairness, and proposed Im-FairAD and Ex-FairAD, two novel fairness-aware anomaly detection methods. Considering the limitations of existing fairness evaluation metrics used in previous works, we propose a novel threshold-free metrics ADPD, which provides a holistic view for evaluating the fairness of methods. 

Empirical results on real-world datasets indicated that the proposed two methods achieve a better trade-off between detection accuracy and fairness than baselines. In most cases, Im-FairAD has better performance than Ex-FairAD. Moreover, we analyzed the reason why fairness can be ensured on the abnormal samples of the test set, although the model training process does not utilize any anomalous samples.

\bibliography{neurips_2025.bib}
\bibliographystyle{plainnat}

%%%%%%%%%%%%%%%%%%%%%%%%%%%%%%%%%%%%%%%%%%%%%%%%%%%%%%%%%%%%

\appendix

\section{Proof for Claim and Propositions}

\subsection{Proof for Claim 1}

\label{proof-claim1}
\begin{proof}
    According to \textbf{Definition}~\ref{def_DP_mc}, to achieve demographic parity, we need to guarantee $\mathbb{P}[f(\mathbf{x}) = \hat{{y}}~|~S=s] = \mathbb{P}[f(\mathbf{x}) = \hat{{y}}]$ for all $s$ and $\hat{y}$. Therefore, we have \\
    \begin{equation*}
        \begin{aligned}
            \mathbb{P}[f(\mathbf{x}) = \hat{{y}}]&=~\mathbb{P}[f(\mathbf{x}) = \hat{y}~|~y=0] \times \mathbb{P}[y=0]\\
            &+~\mathbb{P}[f(\mathbf{x}) = \hat{y}~|~y=1] \times \mathbb{P}[y=1]
        \end{aligned}
    \end{equation*}
    However, only normal data ($y=0$) can be accessed in unsupervised anomaly detection, which means that we can only obtain $\mathbb{P}[f(\mathbf{x}) = \hat{y}~|~y=0] \times \mathbb{P}[y=0]$ in such scenario. Therefore, demographic parity cannot be guaranteed in unsupervised anomaly detection.
    
    And according to \textbf{Definition}~\ref{def_EO_mc}, for achieving equal opportunity, we need to guarantee $\mathbb{P}[\hat{y}=1~|~S=s, y=1] = \mathbb{P}[\hat{y}=1~|~y=1]$ for all $s$. Obviously, we cannot obtain $\mathbb{P}[\hat{y}=1~|~S=s, y=1]$ and $\mathbb{P}[\hat{y}=1~|~y=1]$ in unsupervised anomaly detection if without any additional assumptions. Therefore, equal opportunity cannot be guaranteed in unsupervised anomaly detection.
\end{proof}

\subsection{Proof for Proposition 1}
\label{proof-prop1}
We split $\mathcal{X}$ into different protected groups $\mathcal{X}_{S=s_i}$ according to the values of a sensitive attribute $S$. Therefore, a fair learning on $\mathcal{X}$ is to ensure
\begin{equation}\label{eq-1}
\begin{aligned}
     &\mathbb{P}[\hat{y} ~|~\mathbf{x} \in \mathcal{X}_{S=s_i},\mathcal{T}^*(\mathbf{x}_{\pi_1})\leq \tilde{y} \leq\mathcal{T}^*(\mathbf{x}_{\pi_n})]\\
     =~&\mathbb{P}[\hat{y}~|~\mathbf{x} \in \mathcal{X}_{S=s_j},\mathcal{T}^*(\mathbf{x}_{\pi_1})\leq \tilde{y} \leq\mathcal{T}^*(\mathbf{x}_{\pi_n})] 
\end{aligned}
\end{equation}

we expect to find a $h$ that map data from different demographic groups into a common target distribution where an effective anomaly score function $\zeta$ exist naturally.

Therefore, we obtain a detector $f =\zeta\circ h$ and $\hat{y} = f(\mathbf{x})=\zeta\circ h(\mathbf{x})$.
Indeed, if $h$ is fair for $S$, we have

\begin{equation}\label{eq-3}
\begin{aligned}
     &\mathbb{P}[h(\mathbf{x}) ~|~\mathbf{x} \in \mathcal{X}_{S=s_i},\mathcal{T}^*(\mathbf{x}_{\pi_1})\leq \tilde{y} \leq\mathcal{T}^*(\mathbf{x}_{\pi_n})]\\
     =~&\mathbb{P}[h(\mathbf{x})~|~\mathbf{x} \in \mathcal{X}_{S=s_j},\mathcal{T}^*(\mathbf{x}_{\pi_1})\leq \tilde{y} \leq\mathcal{T}^*(\mathbf{x}_{\pi_n})].    
\end{aligned}
\end{equation}

\begin{proof}
    Equation \eqref{eq-3} indicates that $h(\mathbf{x})$ is independent from $S$. Therefore, $\zeta(h(\mathbf{x}))$ is independent from $S$, that is
    \begin{equation}
    \begin{aligned}
    &\mathbb{P}[\zeta(h(\mathbf{x})) ~|~\mathbf{x} \in \mathcal{X}_{S=s_i},\mathcal{T}^*(\mathbf{x}_{\pi_1})\leq \tilde{y} \leq\mathcal{T}^*(\mathbf{x}_{\pi_n})]\\
     =~&\mathbb{P}[\zeta(h(\mathbf{x}))~|~\mathbf{x} \in \mathcal{X}_{S=s_j},\mathcal{T}^*(\mathbf{x}_{\pi_1})\leq \tilde{y} \leq\mathcal{T}^*(\mathbf{x}_{\pi_n})] 
    \end{aligned}
    \end{equation}
    where $\zeta(h(\mathbf{x})) = \hat{y}$.
\end{proof}

\subsection{Proof for Proposition 2}
\label{proof-prop2}
% \begin{proposition}
%     If $\sum_{s \in S} \mathcal{M}(\mathcal{P}(\mathcal{D}_{\mathcal{X}_{S=s}}), \mathcal{D}_{\mathbf{z}})=0$, then \eqref{eq-3} is attained.
% \end{proposition}
\begin{proof}
Since $\mathcal{M}(\cdot, \cdot)$ is a distance metric between distributions, we have $\mathcal{M}(\cdot, \cdot) \geq 0$.
Therefore, 
\begin{equation}
\begin{aligned}
        &\sum_{s \in S} \mathcal{M}(\mathcal{P}(\mathcal{D}_{\mathcal{X}_{S=s}}), \mathcal{D}_{\mathbf{z}})=0\\
        \Rightarrow~& \mathcal{M}(\mathcal{P}(\mathcal{D}_{\mathcal{X}_{S=s_i}}), \mathcal{D}_{\mathbf{z}})= 0, ~\forall i\\
        \Rightarrow~& \mathcal{P}(\mathcal{D}_{\mathcal{X}_{S=s_i}}) =\mathcal{D}_{\mathbf{z}},~\forall i\\
        \Rightarrow~& \mathcal{P}(\mathcal{D}_{\mathcal{X}_{S=s_i}}) = \mathcal{P}(\mathcal{D}_{\mathcal{X}_{S=s_j}}),~\forall i,j.
\end{aligned}
\end{equation}
It follows that
\begin{equation}
\begin{aligned}
            \mathbb{P}[h(\mathbf{x}) \in \mathcal{P}(\mathcal{D}_{\mathcal{X}_{S=s_i}})] = \mathbb{P}[h(\mathbf{x}) \in \mathcal{P}(\mathcal{D}_{\mathcal{X}_{S=s_j}})]
\end{aligned}
\end{equation}
holds for any $i,j$. Then 
\begin{equation}
           \mathbb{P}[h(\mathbf{x}) | \mathbf{x} \in \mathcal{X}_{S=s_i}] = \mathbb{P}[h(\mathbf{x}) | \mathbf{x} \in \mathcal{X}_{S=s_j}],
\end{equation}
for any $i,j$.
% And, according to \textbf{Assumption 1}, for any $\mathbf{x} \in \mathcal{X}$ satisfies $g(\mathbf{x}_{\pi_1})\leq\mathcal{T}^*(\mathbf{x})\leq\mathcal{T}^*(\mathbf{x}_{\pi_n})$, therefore, we obtain
This means $h(\mathbf{x})$ is independent from $S$. We obtain
\begin{equation}
\begin{aligned}
    &\mathbb{P}[h(\mathbf{x}) ~|~\mathbf{x} \in \mathcal{X}_{S=s_i},\mathcal{T}^*(\mathbf{x}_{\pi_1})\leq \tilde{y} \leq\mathcal{T}^*(\mathbf{x}_{\pi_n})]\\
     =~&\mathbb{P}[h(\mathbf{x})~|~\mathbf{x} \in \mathcal{X}_{S=s_j},\mathcal{T}^*(\mathbf{x}_{\pi_1})\leq \tilde{y} \leq\mathcal{T}^*(\mathbf{x}_{\pi_n})].
\end{aligned}
\end{equation}
\end{proof}

\section{Experimental Settings}

\subsection{Datasets and Baselines}
\label{data-baselines}
The statistics of all datasets are provided in Table~\ref{data-info}. The details of each dataset are as follows:

\begin{table*}[h!]
    \centering
    \caption{Statistics of datasets.}
    \label{data-info}
    \resizebox{\textwidth}{!}{\begin{tabular}{l|c|c|c|c|c}
    \hline
    Dataset & Type & Dimension & Sensitive Variable & Normal Set & Abnormal Set \\
    \hline 
    Adult & tabular & 14 & gender   & income $\leq$50K & income $>$ 50K\\
    \hline
    COMPAS & tabular & 8 & race &  no recidivism within 2 years & recidivism within 2 years\\
    \hline
    Credit & tabular & 23 & age  & no default payment next month & default payment next month\\
    \hline
    Titanic & tabular & 7 & gender  & no survived & survived\\
    \hline
    SP & tabular & 32 & gender & general final grades & extreme final grades  \\
    \hline
    CelebA & image & 64$\times$64$\times$3 & gender  & attractive face & plain face\\
    \hline
    \end{tabular}}
\end{table*}

\begin{itemize}

    \item{\textbf{Adult}\footnote{https://archive.ics.uci.edu/dataset/2/adult}~\citep{misc_adult_2} The dataset is from the 1994 Census Income database contains 48,842 samples with 14 attributes, and gender (male or female) is selected as the sensitive attribute. Following the previous work~\citep{han2023achieving}, we removed the samples with missing values.}
    \item{\textbf{Credit}\footnote{https://archive.ics.uci.edu/dataset/350/default+of+credit+card+clients}~\citep{default_of_credit_card_clients_350} The dataset is about customers' default payments in Taiwan and contains 30,000 samples with 23 attributes. Age is selected as the sensitive attribute where one group includes people from 30 to 60 years of age and the other is other age groups.}
    
    \item{\textbf{Compas}\footnote{https://github.com/propublica/compas-analysis}~\citep{larson2016we} The dataset contains 7,214 samples with 52 attributes. Following previous works~\citep{larson2016we,han2023achieving}, we only selected African-American and Caucasian individuals, yielding 5278 clean samples with 8 attributes. The sensitive attribute is race (African-American and Caucasian).}
    
    \item{\textbf{CelebA}\footnote{https://mmlab.ie.cuhk.edu.hk/projects/CelebA.html}~\citep{liu2015faceattributes} The dataset contains 202,599 color face images. Following~\citep{zhang2021towards}, we resized all images to 64 $\times$ 64. Gender (male or female) is selected as the sensitive attribute.} 
    
    \item{\textbf{Titanic}\footnote{https://www.kaggle.com/c/titanic/data}~\citep{titanic} The dataset is from a kaggle competition to predicting survival on the Titanic. We removed the samples with missing values and obtained 712 clean samples with 8 attributes. Gender (male and female) is selected as the sensitive attribute.}
    \item{\textbf{SP}\footnote{https://archive.ics.uci.edu/dataset/320/student+performance}~\citep{student_performance_320} The dataset is about student achievement in secondary education of two Portuguese schools and contains 649 samples with 32 attributes. The data attributes include student grades, demographic, social and school related features. Gender (male and female) is selected as the sensitive attribute.}
\end{itemize}

We used LOF~\citep{Breunig_Kriegel_Ng_Sander_2000}, Deep SVDD~\citep{Ruff_Vandermeulen_Goernitz_Deecke_Siddiqui_Binder_Müller_Kloft_2018} as standard (fairness-unaware) UAD baselines, FairOD~\citep{Shekhar_Shah_Akoglu_2021}, Deep Fair SVDD~\citep{zhang2021towards}, and CFAD~\citep{han2023achieving} as end-to-end fairness-aware UAD baselines. We utilized FRL technique FarconVAE~\citep{oh2022learning} to construct ``FarconVAE+LOF'' and ``FarconVAE+DeepSVDD'' as two-stage pipelines.

The implementations of LOF and Deep SVDD are built using the PyOD\footnote{https://github.com/yzhao062/pyod} library~\citep{zhao2019pyod}. FarconVAE\footnote{https://github.com/changdaeoh/FarconVAE}, FairOD\footnote{https://github.com/Shubhranshu-Shekhar/fairOD} and CFAD\footnote{https://github.com/hanxiao0607/CFAD} are based on official codes and the hyperparameters are fine-tuned according to the suggestions from original papers. In light of the unavailable implementation for Deep Fair SVDD, we reproduce the code following the pseudo-code provided in the original paper~\citep{zhang2021towards}. 

\subsection{Implementation Details}
\label{exp-settings}

\paragraph{Neural Network Architectures} For the tabular datasets Adult, COMPAS, Credit, Titanic and SP, the neural networks for all methods are Multi-Layer Perceptrons (MLP). For the image dataset CelebA, the neural networks used in all methods are Convolutional Neural Networks (CNN). We use Adam~\citep{kingma2015adam} as the optimizer, and set coefficient $\alpha$ of entropy regularization term in the Sinkhorn distance to $0.1$ in all experiments. 

\paragraph{Data Splitting} For fairness problems in unsupervised anomaly detection, the proportion of sample size across different demographic groups heavily influences the results~\citep{meissen2023predictable, wu2024fair}. Therefore, in this work, we set balanced splitting and skewed splitting. The detailed splits for the training set and test set are provided in Tables~\ref{tab-split-balanced} and \ref{tab-split-skewed}. Note that there is no balanced splitting for Titanic and SP due to the limitations of extremely uneven samples size across different demographic groups for Titanic and insufficient sample size for SP.

\begin{table}[h!]
    \centering
    \caption{Balanced data splitting on Adult, COMPAS, CelebA and Credit. PV denotes the protected variable. `Nor.' and `Abnor.' denote normal sample and abnormal sample, respectively.}
    \label{tab-split-balanced}
    \resizebox{0.85\textwidth}{!}{
    \begin{tabular}{|c|c|c|c|c|}
    \hline
    \multirow{2}{*}{Dataset (PV)} & \multirow{2}{*}{Attribute Value} & Training Set & \multicolumn{2}{c|}{Test Set} \\
    \cline{3-5}
    & & Size (Nor.) & Size (Nor.) & Size (Abnor.)\\
    \hline
    \multirow{2}{*}{Adult (Gender)} &  Male & 6000 & 1000 & 1000 \\
     &  Female & 6000 & 1000 & 1000 \\
    \hline
    \multirow{2}{*}{COMPAS (Race)} &  African-American & 1000 & 280 & 280 \\
     &  Female & 1000 & 280 & 280 \\
    \hline
    \multirow{2}{*}{CelebA (Gender)} &  Male & 8000 & 4000 & 4000 \\
     &  Female & 8000 & 4000 & 4000 \\
    \hline
    \multirow{2}{*}{Credit (Age)} &  [30, 60] & 5000 & 2000 & 2000 \\
     &  Others & 5000 & 2000 & 2000 \\
    \hline
    \end{tabular}}
\end{table}

\begin{table}[h!]
    \centering
    \caption{Skewed data splitting on all datasets. PV denotes the protected variable. `Nor.' and `Abnor.' denote normal sample and abnormal sample, respectively.}
    \label{tab-split-skewed}
    \resizebox{0.85\textwidth}{!}{
    \begin{tabular}{|c|c|c|c|c|}
    \hline
    \multirow{2}{*}{Dataset (PV)} & \multirow{2}{*}{Attribute Value} & Training Set & \multicolumn{2}{c|}{Test Set} \\
    \cline{3-5}
    & & Size (Nor.) & Size (Nor.) & Size (Abnor.)\\
    \hline
    \multirow{2}{*}{Adult (Gender)} &  Male & 8000 & 4000 & 4000 \\
     &  Female & 2000 & 1000 & 1000 \\
    \hline
    \multirow{2}{*}{COMPAS (Race)} &  African-American & 800 & 400 & 400 \\
     &  Female & 200 & 100 & 100 \\
    \hline
    \multirow{2}{*}{CelebA (Gender)} &  Male & 8000 & 4000 & 4000 \\
     &  Female & 2000 & 1000 & 1000 \\
    \hline
    \multirow{2}{*}{Credit (Age)} &  [30, 60] & 8000 & 4000 & 4000 \\
     &  Others & 2000 & 1000 & 1000 \\
        \hline
    \multirow{2}{*}{Titanic (Gender)} &  Male & 330 & 30 & 30 \\
     &  Female & 32 & 30 & 30 \\
    \hline
    \multirow{2}{*}{SP (Gender)} &  Male & 135 & 26 & 30 \\
     &  Female & 148 & 26 & 30 \\
    \hline
    \end{tabular}}
\end{table}

\paragraph{Evaluation Metrics} Following previous works such as \citep{lahoti2020fairness,buyl2022optimal}, we use the AUC (Area Under the Receiver Operating Characteristic curve) score and F1 to evaluate the detection accuracy of all compared methods. We use the proposed ADPD (threshold-free) and \textit{fairness ratio} (threshold-dependent)~\citep{zhang2021towards} to evaluate the fairness of unsupervised anomaly detection. Note that the threshold is determined by percentile ($p$) of anomaly score on the training data and we set $p=\{0.9, 0.95\}$ for all baselines.

ALL experiments were conducted on 20 Cores Intel(R) Xeon(R) Gold 6248 CPU with one NVIDIA Tesla V100 GPU, CUDA 12.0. We run each experiment five times and report the average results with standard variance.

\subsection{Threshold-Free Fairness Metrics}
\label{dis-adpd-fr}

When evaluating the fairness of anomaly detection, existing work such as~\citep{zhang2021towards,Shekhar_Shah_Akoglu_2021} usually uses a \textit{fairness ratio}
\begin{equation}
    r := \min \left(\tfrac{\mathbb{P}(\text{Score}(\mathcal{X})>t | S=s_i)}{\mathbb{P}(\text{Score}(\mathcal{X})>t | S=s_j)}, \tfrac{\mathbb{P}(\text{Score}(\mathcal{X})>t | S=s_j)}{\mathbb{P}(\text{Score}(\mathcal{X})>t | S=s_i)}\right)
    \label{eq7}
\end{equation}
where $\text{Score}(\cdot)$ denotes the anomaly score, $t$ is a threshold and $S$ denotes a protected variable. The metric has two limitations.
First, ~\eqref{eq7} is sensitive to the selection of threshold $t$. 
Second, ~\eqref{eq7} will not work (undefined cases) when $\mathbb{P}(\text{Score}(\mathbf{x})>t | S=s_j) = 0$ or $\mathbb{P}(\text{Score}(\mathbf{x})>t | S=s_i) = 0$, where $r$ is always zero. 
To overcome the two limitations, in this paper, we propose a new fairness metric called \textit{Average Demographic Parity Difference} (ADPD):
\begin{equation}
% \begin{aligned}
%     \text{ADPD} := \frac{1}{n\cdot \lfloor \frac{|\mathcal{S}|^2}{4} \rfloor}\sum_{k=1}^{n} \sum_{i\neq j}^{|\mathcal{S}|}\Big\vert & \mathbb{P}(\text{Score}(\mathcal{X}) > t_k | S=s_i ) \\
%     - & \mathbb{P}(\text{Score}(\mathcal{X}) > t_k | S=s_j) \Big\vert
%     \label{eq7-1}
% \end{aligned}
% \begin{aligned}
    \text{ADPD} := \frac{1}{n}\sum_{k=1}^{n} \Big\vert  \mathbb{P}(\text{Score}(\mathcal{X}) > t_k | S=s_i ) 
    -  \mathbb{P}(\text{Score}(\mathcal{X}) > t_k | S=s_j) \Big\vert
    \label{eq7-1}
% \end{aligned}
\end{equation}
where $t_k \in \text{Score}(\mathcal{X})$ denotes the anomaly score of single sample. In our proposed methods, $t_k = \Vert h_{\phi^\ast}(\mathbf{x}_k)\Vert$. ADPD is a threshold-free metric (such as AUC) for measuring demographic parity. The range of $\text{ADPD}$ is $[0, 1)$ and a smaller ADPD means a higher fairness. Overall, due to threshold-independent, ADPD provides a holistic view of the model's fairness by evaluating the performance of an AD method across all possible thresholds. Although we introduce a novel threshold-free metric for fairness measure, this is not to imply that the threshold-dependent metrics are useless.

The ADPD evaluates the performance of a model across all possible thresholds, providing a holistic view of the model's fairness. In contrast, the \textit{fairness ratio} is calculated based on a specific threshold and exhibits high sensitivity to threshold selection (See Table~\ref{tab-fr-f1-cacc} and Table~\ref{tab-fr-f1-ts}). When prior knowledge (e.g., anomaly prevalence, operational constraints) is unavailable, ADPD is preferable as it mitigates bias introduced by arbitrary threshold choices, ensuring equitable performance comparisons. On the other hand, if domain-specific priors (e.g., training set contamination rate) are available, threshold-dependent metrics like the \textit{fairness ratio} may align better with real requirements.

\section{Numerical Results and More Visualization}
\label{more-results}

\subsection{Results of ADPD and AUC on all datasets}
\label{res-adpd-auc}
\begin{table}[h!]
    \centering
    \caption{Results of ADPD and AUC on COMPAS, Adult, Credit and CelebA. Note that the baselines, FairOD and CFAD are tailored to tabular data. In the official code of FarconVAE, there is no support for image data.}
    \label{tab-adpd-auc-cacc}
    \resizebox{\columnwidth}{!}{
    \begin{tabular}{l|c|c|c|c|c|c}
    \toprule
    \multirow{3}{*}{Methods} & \multicolumn{3}{c|}{Balanced Splitting} & \multicolumn{3}{c}{Skewed Splitting} \\
    \cline{2-7}
    & \multirow{2}{*}[-0.85ex]{AUC(\%)~$\uparrow$} & \multicolumn{2}{c|}{ADPD(\%)~$\downarrow$} & \multirow{2}{*}[-0.85ex]{AUC(\%)~$\uparrow$} & \multicolumn{2}{c}{ADPD(\%)~$\downarrow$} \\
    \cline{3-4} \cline{6-7}
    & & normal & all & & normal & all \\
    \midrule
    \multicolumn{7}{c}{\cellcolor{gray!40}COMPAS} \\
    \midrule
    LOF & 59.23(0.00) & 12.27(0.00) & 10.59(0.00) & 57.25(0.00) & 9.33(0.00) & 9.53(0.00)\\
    Deep SVDD & 57.58(2.30) & 24.33(7.87) & 21.07(8.72) & 60.24(3.08) &  31.72(7.39) & 30.18(5.32)\\
    FarconVAE+LOF & 48.72(0.00) & 4.56(0.00) & 6.56(0.00) & 50.10(0.00) & 7.25(0.00) & \textbf{4.12}(0.00)\\
    FarconVAE+Deep SVDD & 50.48(1.67) & \textbf{3.99}(1.51) & 4.77(0.72) & 48.83(0.75) & 8.50(1.25) & 7.03(1.10) \\
    FairOD & 61.05(1.05) & 5.39(2.11) & 5.53(1.50) & 58.86(4.75) &  \textbf{5.43}(0.60) &  \textbf{5.32}(3.24) \\
    Deep Fair SVDD & 62.71(0.76) & 10.25(2.40) &  8.25(1.46) &  61.05(1.07) &  10.58(3.23) &  11.04(3.13) \\ 
    CFAD & 62.27(0.68) & 12.21(4.57) &  10.94(4.40) &  60.87(1.99) & 23.81(3.71) &  17.39(2.20)\\
    \midrule
    Ex-FairAD (Ours) & \makecell[c]{\textbf{63.11}}(2.67) & \makecell[c]{{4.38}}(0.43) & \makecell[c]{\textbf{4.54}}(1.38) & \makecell[c]{\textbf{66.44}}(3.62) & \makecell[c]{5.44}(1.05) & \makecell[c]{7.39}(1.18)\\
    Im-FairAD (Ours) & \makecell[c]{\textbf{63.89}}(3.45) & \makecell[c]{\textbf{4.16}}(2.55) & \makecell[c]{\textbf{4.76}}(2.37) & \makecell[c]{\textbf{66.58}}(2.91) & \makecell[c]{\textbf{4.90}}(1.00) & \makecell[c]{5.43}(0.49)\\
    \midrule
    \multicolumn{7}{c}{\cellcolor{gray!40}Adult} \\
    \midrule
    LOF & 58.09(0.00) & 26.46(0.00) & 25.58(0.00) & 59.14(0.00) & 25.19(0.00) & 25.34(0.00)\\
    Deep SVDD & 60.47(3.59) & 7.09(1.92) & 7.35(1.62) & 61.04(3.07) & 19.43(5.40) & 17.32(2.90)\\
    FarconVAE+LOF & 53.92(0.00) & 5.00(0.00) & 5.73(0.00) & 50.56(0.00) & 2.16(0.00) & 3.36(0.00)\\
    FarconVAE+Deep SVDD & 55.05(2.81) & 4.65(3.01) & 5.76(3.24) & 51.21(2.61) & 5.65(5.84) & 5.81(7.09)\\
    FairOD & 70.63(0.36) & 11.80(4.62) & 9.26(1.45) & 58.30(4.18) & 6.34(0.53) &  14.35(5.97) \\
    Deep Fair SVDD & 62.45(0.65) &  \textbf{2.18}(0.84) &  3.01(2.27) & 60.22(0.14) & 6.86(1.85) &  4.14(3.74) \\ 
    CFAD & 66.89(1.12) & 6.07(2.95) &  14.69(3.58) & 60.28(1.09) & 23.42(5.43) & 30.41(3.64)\\
    \midrule
    Ex-FairAD (Ours) & \makecell[c]{\textbf{73.49}}(4.41) & \makecell[c]{2.92}(0.81) & \makecell[c]{\textbf{2.86}}(1.09) & \makecell[c]{\textbf{69.56}}(3.45) & \makecell[c]{\textbf{1.52}}(0.44) & \makecell[c]{\textbf{1.79}}(0.76)\\
    Im-FairAD (Ours) & \makecell[c]{\textbf{72.05}}(1.81) & \makecell[c]{\textbf{1.61}}(0.61) & \makecell[c]{\textbf{2.13}}(1.17) & \makecell[c]{\textbf{69.34}}(1.82) & \makecell[c]{\textbf{1.92}}(1.05) & \makecell[c]{\textbf{1.70}}(0.86)\\
    \midrule
    \multicolumn{7}{c}{\cellcolor{gray!40}Credit} \\
    \midrule
    LOF & 50.09(0.00) & 6.28(0.00) & 3.19(0.00) & 48.98(0.00) & 2.14(0.00) & 3.17(0.00)\\
    Deep SVDD & 52.34(1.98) & 9.84(4.56) & 8.64(3.26) & 54.82(1.97) & 7.49(1.33) & 6.95(2.55)\\
    FarconVAE+LOF & 49.86(0.00) & 2.21(0.00) & 2.24(0.00) & 52.93(0.00) & 2.65(0.00) & \textbf{0.80}(0.00)\\
    FarconVAE+Deep SVDD & 51.07(0.58) & \textbf{1.03}(0.45) & \textbf{1.01}(0.32) & 50.96(0.25) & \textbf{1.49}(0.70) & \textbf{0.88}(0.66)\\
    FairOD & 53.93(1.09) & 7.00(1.33) & 7.21(1.42) & 60.39(0.55) & 8.79(0.73) &  7.36(0.50) \\
    Deep Fair SVDD & 56.01(0.69) & 4.62(1.11) &  4.54(1.22) & 56.67(0.42) & 6.24(1.88) &  5.84(2.00) \\ 
    CFAD & 56.96(0.55) & 5.89(0.75) & 4.48(0.66) & 57.21(0.47) & 6.30(2.25) & 5.76(1.94)\\
    \midrule
    Ex-FairAD (Ours) & \makecell[c]{\textbf{64.96}}(1.63) & \makecell[c]{2.95}(0.83) & \makecell[c]{\textbf{1.92}}(0.39) & \makecell[c]{\textbf{63.85}}(1.95) & \makecell[c]{2.71}(0.85) & \makecell[c]{1.57}(0.87)\\
    Im-FairAD (Ours) & \makecell[c]{\textbf{63.70}}(1.59) & \makecell[c]{\textbf{2.20}}(0.86) & \makecell[c]{1.97}(0.66) & \makecell[c]{\textbf{65.13}}(1.79) & \makecell[c]{\textbf{2.34}}(0.59) & \makecell[c]{1.33}(0.24)\\
    \midrule
    \multicolumn{7}{c}{\cellcolor{gray!40}CelebA} \\
    \midrule
    Deep SVDD & \textbf{62.77}(0.35) & 10.44(1.66) & 10.47(1.55) & 60.80(0.68) & 9.82(0.44) & 8.24(0.93)\\
    Deep Fair SVDD & 57.97(0.72) & 7.73(0.99) & 4.79(0.35) & 59.95(0.80) & \textbf{1.52}(1.21) & \textbf{3.24}(1.81) \\ 
    \midrule
    Ex-FairAD (Ours) & \makecell[c]{\textbf{60.68}}(0.72) & \makecell[c]{\textbf{1.06}}(0.47) & \makecell[c]{\textbf{3.80}}(0.83) & \makecell[c]{\textbf{63.07}}(0.71) & \makecell[c]{\textbf{1.91}}(0.38) & \makecell[c]{4.87}(0.60) \\
    Im-FairAD (Ours) & \makecell[c]{58.07}(0.87) & \makecell[c]{\textbf{2.37}}(0.98) & \makecell[c]{\textbf{2.46}}(0.45) & \makecell[c]{\textbf{61.72}}(0.62) & \makecell[c]{2.73}(0.86) & \makecell[c]{\textbf{2.93}}(0.54)\\
    \bottomrule
    \end{tabular}
    }
\end{table}

The results of ADPD and AUC on all datasets are provided in Table~\ref{tab-adpd-auc-cacc} and Table~\ref{tab-adpd-auc-ts}, where the best two results in each case are marked in \textbf{bold}, `normal' means the score is computed only on normal samples of the test set, and `all' means the score is computed on all samples from test set. From these two tables, we have the following observations:

\begin{itemize}

    \item Im-FairAD and Ex-FairAD both achieve better detection accuracy (AUC) compared with all baselines not only on normal data but also on entire test set in most cases, while maintaining comparable or even better fairness (ADPD) \textbf{[Q1]}.

    \item In the transition from a balanced splitting to a skewed splitting, group fairness, as measured by ADPD, exhibits significantly larger values in most cases, which indicates the fairness of existing fairness-aware AD methods is easily affected by sample proportion from different sensitive groups and skewed data for different demographic groups poses a more intractable fairness problem than balanced data. In contrast, the fluctuation of ADPD is slight on our proposed methods \textbf{[Q2]}.
\end{itemize}

\begin{table}[h!]
    \centering
    \caption{Results of ADPD and AUC on Titanic and SP with gender as the sensitive attribute. Note that there is no balanced splitting for Titanic and SP due to the limitations of extremely uneven samples size across different demographic groups for Titanic and insufficient sample size for SP.}
    \label{tab-adpd-auc-ts}
    \resizebox{\columnwidth}{!}{
    \begin{tabular}{l|c|c|c|c|c|c}
    \toprule
    \multirow{3}{*}{Methods} & \multicolumn{3}{c|}{Titanic} & \multicolumn{3}{c}{SP} \\
    \cline{2-7}
    & \multirow{2}{*}[-0.85ex]{AUC(\%)~$\uparrow$} & \multicolumn{2}{c|}{ADPD(\%)~$\downarrow$} & \multirow{2}{*}[-0.85ex]{AUC(\%)~$\uparrow$} & \multicolumn{2}{c}{ADPD(\%)~$\downarrow$} \\
    \cline{3-4} \cline{6-7}
    & & normal & all & & normal & all \\
    \midrule
    LOF & 53.88(0.00) & 23.16(0.00) & 19.81(0.00) & 63.55(0.00) & 9.82(0.00) & 6.03(0.00)\\
    Deep SVDD & 53.21(1.98) & 26.02(11.29) & 26.07(11.95) & 68.37(1.11) & 12.97(4.03) & 12.52(2.44)\\
    FarconVAE+LOF & 48.52(0.00) & 7.97(0.00) & 6.58(0.00) & 57.30(0.00) & \textbf{4.54}(0.00) & 8.94(0.00)\\
    FarconVAE+Deep SVDD & 50.04(2.97) & 7.60(1.89) & \textbf{3.99}(0.85) & 55.00(3.17) & 9.53(3.44) & 5.98(1.13)\\
    FairOD &  46.04(0.20) & 14.67(0.30) & 6.38(0.16) & 42.04(0.44) & 9.70(0.51) & 10.90(0.31)\\
    Deep Fair SVDD & 51.94(0.99) & 20.67(3.08) & 21.24(6.35) & 61.90(1.20) & 10.57(3.04) & 7.24(2.12)\\ 
    CFAD & 55.63(0.59) & 26.77(7.00) & 23.02(2.10) & 69.23(1.34) & 9.26(1.24) & \textbf{4.53}(0.28)\\
    \midrule
    Ex-FairAD (Ours) & \textbf{64.87}(1.48) & \textbf{6.35}(2.12) & 5.75(1.00) & \textbf{74.16}(3.09) & 6.00(1.32) & 5.84(1.67)\\
    Im-FairAD (Ours) & \textbf{64.02}(3.62) & \textbf{7.14}(2.61) & \textbf{3.82}(1.77) & \textbf{75.97}(4.32) & \textbf{5.84}(1.29) & \textbf{5.62}(1.95)\\ 
    \bottomrule
    \end{tabular}
    }
\end{table}

\begin{figure*}[h!]
    \centering
    \includegraphics[width=0.98\textwidth]{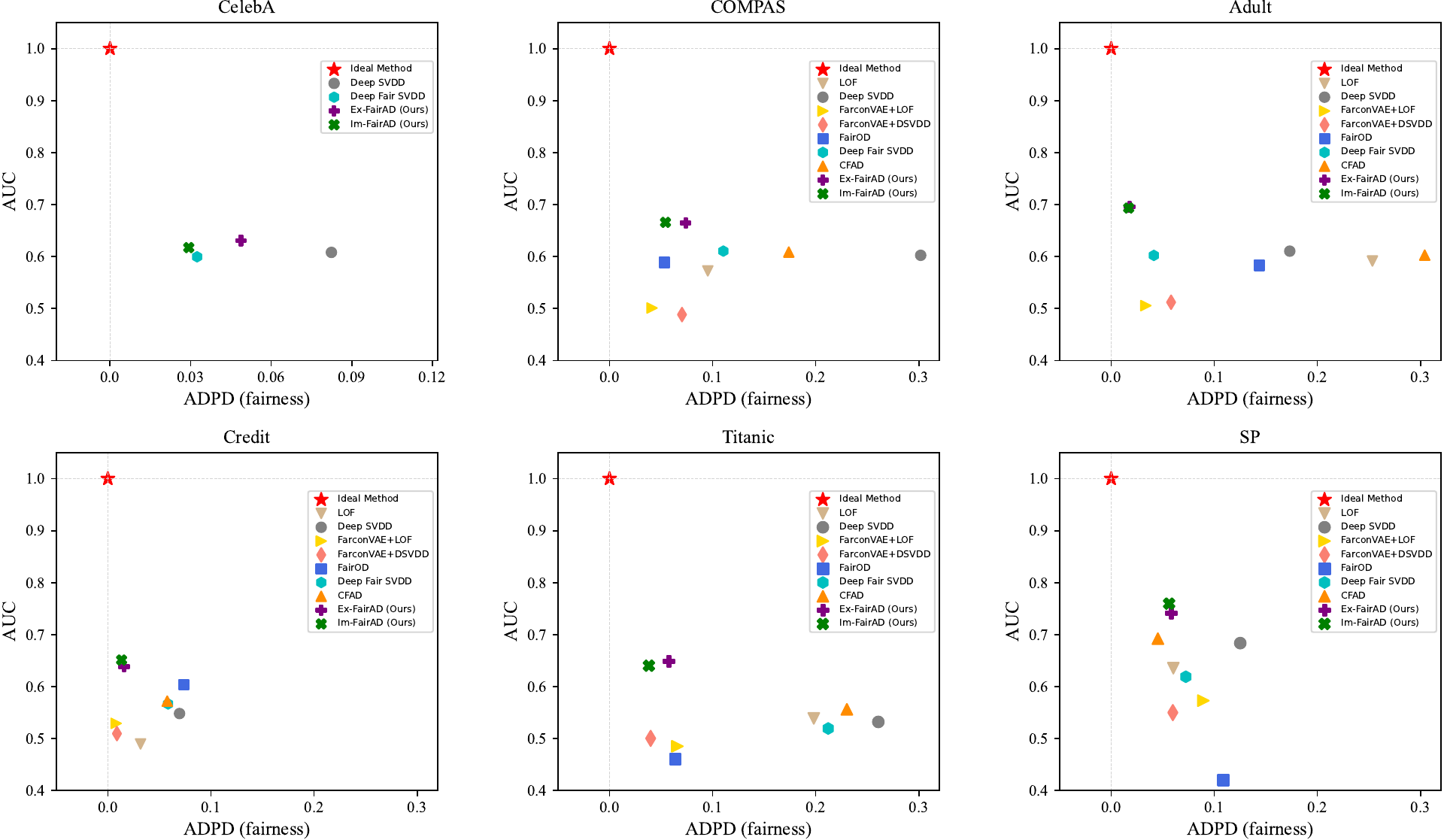}
    \caption{Accuracy-fairness trade-off on skewed splitting.}
    \label{fig-auc-vs-adpd-skewed}
\end{figure*}

\begin{figure}[h!]
    \centering
    \includegraphics[width=0.48\textwidth]{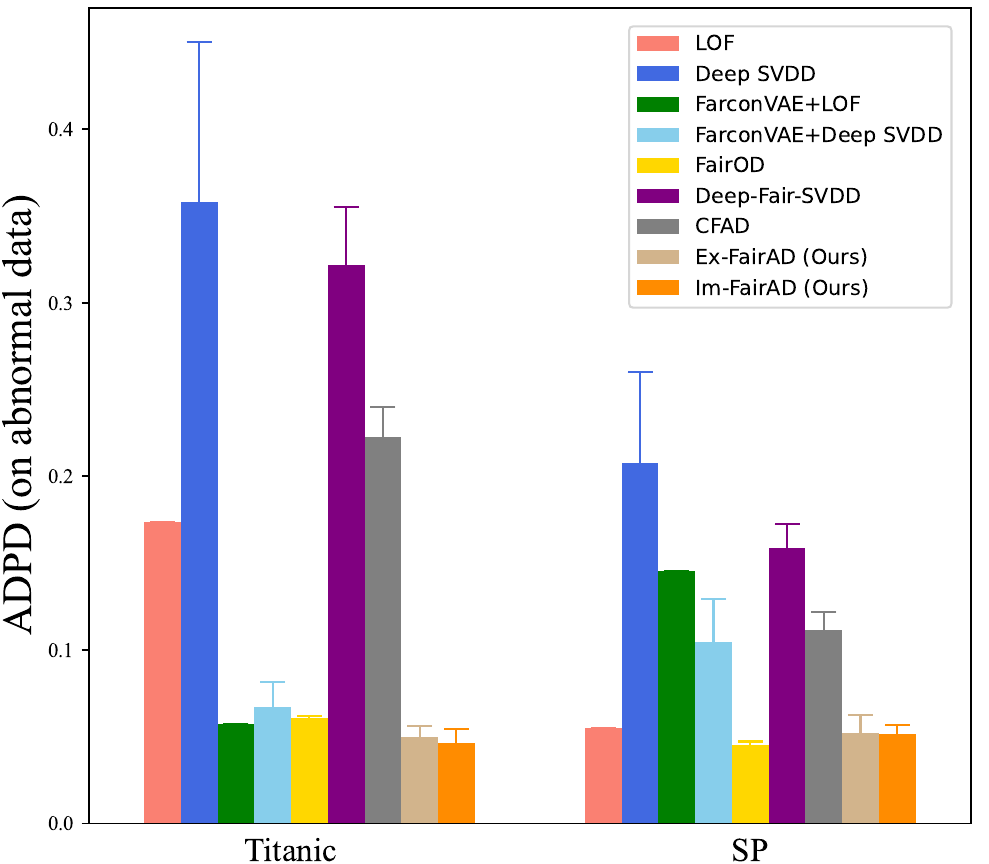}
    \caption{Fairness of all baselines on abnormal data from test set.}
    \label{fig-abnormal-adpd-cr-ti-sp}
\end{figure}

The visualization between detection accuracy (AUC) and fairness (ADPD) on skewed splitting is shown in Figure~\ref{fig-auc-vs-adpd-skewed}. The ADPD of abnormal data on Titanic and SP is shown in Figure~\ref{fig-abnormal-adpd-cr-ti-sp}.

\subsection{Results of \textit{fairness ratio} and F1 on all datasets}
\label{res-fr-f1}

Furthermore, the results of F1-score and \textit{fairness ratio} on all datasets are provided in Table~\ref{tab-fr-f1-cacc} and Table~\ref{tab-fr-f1-ts}, where \textit{fairness ratio} is highly sensitive to different thresholds $p=\{0.90, 0.95\}$.

\begin{table}[h!]
    \centering
    \caption{Results of F1-score and \textit{fairness ratio} on COMPAS, Adult, Credit and CelebA. Note that the baselines, FairOD and CFAD are tailored to tabular data. In the official code of FarconVAE, there is no support for image data.}
    \label{tab-fr-f1-cacc}
    \resizebox{\columnwidth}{!}{
    \begin{tabular}{l|c|c|c|c|c|c}
    \toprule
    \multirow{3}{*}{Methods} & \multicolumn{3}{c|}{Threshold=0.90} & \multicolumn{3}{c}{Threshold=0.95} \\
    \cline{2-7}
    & \multirow{2}{*}{F1(\%)~$\uparrow$} & \multicolumn{2}{c|}{Fairness ratio~$\uparrow$} & \multirow{2}{*}{F1(\%)~$\uparrow$} & \multicolumn{2}{c}{Fairness ratio~$\uparrow$} \\
    \cline{3-4}\cline{6-7}
    & & normal & all & & normal & all \\
    \midrule
    \multicolumn{7}{c}{\cellcolor{gray!40}COMPAS} \\
    \midrule
    LOF & 29.85(0.00) & 0.27(0.00) & 0.35(0.00) & 19.16(0.00) & 0.34(0.00) & 0.36(0.00)\\
    Deep SVDD & 31.30(5.17) & 0.49(0.18) & 0.58(0.18) & 20.67(5.09) & 0.39(0.18) & 0.53(0.30)\\
    FarconVAE+LOF & 14.28(0.00) & 0.69(0.00) & \textbf{0.77}(0.00) & 8.11(0.00) & 0.64(0.00) & 0.55(0.00)\\
    FarconVAE+Deep SVDD & 19.50(0.89) & \textbf{0.74}(0.07) & 0.70(0.14) & 8.58(0.65) & \textbf{0.68}(0.22) & \textbf{0.75}(0.23)\\
    FairOD & 17.81(0.11) & 0.70(0.04) & \textbf{0.79}(0.01) & 14.54(0.66) & \textbf{0.87}(0.01) & \textbf{0.88}(0.06)\\

    Deep Fair SVDD & 42.23(0.81) & 0.46(0.04) & 0.58(0.04) & \textbf{31.54}(0.63) & 0.53(0.09) & 0.56(0.03)\\

    CFAD & 40.67(3.67) & 0.39(0.13) & 0.49(0.08) & 26.95(1.49) & 0.37(0.06) & 0.37(0.03) \\
    \midrule
    Ex-FairAD (Ours) & \textbf{42.30}(5.88) & \textbf{0.71}(0.18) & 0.74(0.15) & 28.07(5.06) & 0.58(0.21) & 0.51(0.07)\\
    Im-FairAD (Ours) & \textbf{47.49}(4.32) & 0.65(0.28) & 0.68(0.09) & 
    \textbf{33.88}(2.75) & 0.57(0.21) & 0.68(0.13) \\
    \midrule
    \multicolumn{7}{c}{\cellcolor{gray!40}Adult} \\
    \midrule
    LOF & 27.45(0.00) & 0.34(0.00) & 0.42(0.00) & 24.58(0.00) & 0.30(0.00) & 0.34(0.00)\\
    Deep SVDD & 44.73(4.72) & 0.54(0.24) & 0.63(0.11) & 26.01(13.02) & 0.48(0.13) & 0.54(0.07)\\
    FarconVAE+LOF & 22.87(0.00) & 0.63(0.00) & 0.62(0.00) & 13.33(0.00) & 0.73(0.00) & 0.58(0.00)\\
    FarconVAE+Deep SVDD & 22.08(8.87) & 0.59(0.10) & 0.73(0.18) & 10.33(4.14) & 0.46(0.27) & 0.48(0.22)\\
    FairOD & 36.87(3.99) & 0.70(0.06) & 0.32(0.04) & 24.82(1.61) & 0.67(0.04) & 0.81(0.05)\\   
    Deep Fair SVDD & 45.03(1.56) & \textbf{0.86}(0.01) & 0.83(0.05) & 39.43(1.35) & \textbf{0.83}(0.08) & \textbf{0.86}(0.06)\\
    CFAD & 48.01(5.51) & 0.76(0.08) & 0.55(0.14) & 35.76(1.60) & 0.77(0.06) & 0.63(0.10)\\
    \midrule
    Ex-FairAD (Ours) & \textbf{52.75}(2.19) & \textbf{0.87}(0.11) & \textbf{0.92}(0.03) & \textbf{48.46}(4.87) & 0.80(0.12) & 0.83(0.07)\\
    Im-FairAD (Ours) & \textbf{56.04}(4.70) & 0.72(0.06) & \textbf{0.84}(0.07) & \textbf{47.79}(1.84) & \textbf{0.90}(0.06) & \textbf{0.92}(0.04)\\
    \midrule
    \multicolumn{7}{c}{\cellcolor{gray!40}Credit} \\
    \midrule
    LOF & 20.11(0.00) & 0.45(0.00) & 0.55(0.00) & 11.78(0.00) & 0.24(0.00) & 0.33(0.00)\\
    Deep SVDD & 23.31(7.75) & 0.76(0.12) & 0.74(0.20) & 16.26(8.71) & 0.72(0.11) & 0.63(0.21)\\
    FarconVAE+LOF & 15.06(0.00) & 0.88(0.00) & 0.88(0.00) & 8.06(0.00) & 0.79(0.00) & 0.83(0.00)\\
    FarconVAE+Deep SVDD & 16.96(1.80) & 0.94(0.02) & 0.95(0.02) & 8.70(1.64) & 0.85(0.12) & 0.87(0.10)\\
    FairOD & 41.00(0.81) & 0.70(0.04) & 0.74(0.02) & 35.18(1.02) & 0.66(0.04) & 0.70(0.07)\\
    Deep Fair SVDD & 13.54(1.79) & \textbf{0.81}(0.06) & 0.73(0.09) & 15.47(0.76) & 0.71(0.03) & 0.70(0.05)\\
    CFAD & 20.43(0.75) & 0.69(0.05) & 0.69(0.04) & 10.53(0.22) & 0.59(0.05) & 0.58(0.02) \\
    \midrule
    Ex-FairAD (Ours) & \textbf{53.70}(2.53) & 0.79(0.03) & \textbf{0.90}(0.04) & \textbf{41.06}(3.79) & \textbf{0.77}(0.06) & \textbf{0.82}(0.10)\\
    Im-FairAD (Ours) & \textbf{54.32}(3.43) & \textbf{0.81}(0.08) & \textbf{0.90}(0.04) & \textbf{44.30}(1.94) & \textbf{0.72}(0.05) & \textbf{0.86}(0.05)\\
    \midrule
    \multicolumn{7}{c}{\cellcolor{gray!40}CelebA} \\
    \midrule
    Deep SVDD & 30.86(1.38) & \textbf{0.91}(0.06) & 0.81(0.05) & 23.08(1.24) & 0.82(0.09) & 0.76(0.03)\\
    Deep Fair SVDD & 14.40(2.8) & 0.58(0.18) & 0.65(0.11) & 16.95(1.32) & 0.63(0.22) & 0.69(0.12)\\
    \midrule
    Ex-FairAD (Ours) & \textbf{50.29}(9.04) & 0.86(0.10) & \textbf{0.84}(0.11) & \textbf{55.40}(10.57) & \textbf{0.83}(0.08) & \textbf{0.88}(0.08)\\
    Im-FairAD (Ours) & \textbf{30.89}(2.34) & \textbf{0.89}(0.08) & \textbf{0.88}(0.07) & \textbf{19.83}(2.88) & \textbf{0.87}(0.04) & \textbf{0.84}(0.07)\\
    \bottomrule
    \end{tabular}}
\end{table}

\begin{table}[h!]
    \centering
    \caption{Results of F1-score and \textit{fairness ratio} on Titanic and SP. Note that `$0^*$' means that undefined case occurs when calculating \textit{fairness ratio}.}
    \label{tab-fr-f1-ts}
    \resizebox{\columnwidth}{!}{
    \begin{tabular}{l|c|c|c|c|c|c}
    \toprule
    \multirow{3}{*}{Methods} & \multicolumn{3}{c|}{Threshold=0.90} & \multicolumn{3}{c}{Threshold=0.95} \\
    \cline{2-7}
    & \multirow{2}{*}{F1(\%)~$\uparrow$} & \multicolumn{2}{c|}{Fairness ratio~$\uparrow$} & \multirow{2}{*}{F1(\%)~$\uparrow$} & \multicolumn{2}{c}{Fairness ratio~$\uparrow$} \\
    \cline{3-4}\cline{6-7}
    & & normal & all & & normal & all \\
    \midrule
    \multicolumn{7}{c}{\cellcolor{gray!40}Titanic} \\
    \midrule
    LOF & 35.55(0.00) & 0.40(0.00) & 0.57(0.00) & 26.82(0.00) & 0.22(0.00) & 0.46(0.00)\\
    Deep SVDD & 31.90(10.57) & 0.44(0.32) & 0.44(0.30) & 16.66(6.47) & 0.68(0.36) & 0.56(0.33)\\
    FarconVAE+LOF & 8.82(0.00) & 0.66(0.00) & 0.6(0.00) & 6.15(0.00) & 0.50(0.00) & 0.25(0.00)\\
    FarconVAE+Deep SVDD & 17.06(5.66) & 0.57(0.21) & 0.69(0.19) & 9.03(4.70) & 0.16(0.21) & 0.28(0.16)\\
    FairOD & 15.55(2.77) & 0.57(0.09) & 0.57(0.04) & 11.42(0.00) & 0.50(0.00) & \textbf{1.00}(0.00)\\
    Deep Fair SVDD & 14.31(2.26) & 0.68(0.30) & 0.66(0.23) & 12.83(1.46) & 0.80(0.24) & 0.40(0.15)\\
    CFAD & 44.54(5.47) & 0.37(0.13) & 0.57(0.24) & 27.73(1.02) & 0.53(0.06) & 0.48(0.05) \\
    \midrule
    Ex-FairAD (Ours) & \textbf{66.66}(0.00) & \textbf{1.00}(0.00) & \textbf{1.00}(0.00) & \textbf{66.82}(0.00) & \textbf{1.00}(0.00) & \textbf{1.00}(0.00)\\

    Im-FairAD (Ours) & \textbf{66.66}(0.00) & \textbf{1.00}(0.00) & \textbf{1.00}(0.00) & \textbf{66.66}(0.00) & \textbf{1.00}(0.00) & \textbf{1.00}(0.00)\\

    \midrule
    \multicolumn{7}{c}{\cellcolor{gray!40}SP} \\
    \midrule
    LOF & 44.44(0.00) & 0.14(0.00) & 0.57(0.00) & 27.02(0.00) & 0.12(0.00) & 0.24(0.00)\\
    Deep SVDD & 22.45(6.56) & 0.38(0.30) & 0.47(0.19) & 20.42(8.47) & 0.59(0.35) & 0.46(0.32)\\
    FarconVAE+LOF & 19.04(0.00) & $0^*$ & 0.65(0.00) & 6.89(0.00) & $0^*$ & $0^*$\\
    FarconVAE+Deep SVDD & 14.41(2.98) & 0.40(0.36) & 0.59(0.26) & 12.90(3.70) & $0^*$ & $0^*$\\
    FairOD & 11.72(0.13) & \textbf{0.88}(0.02) & 0.68(0.06) & 13.65(1.41) & \textbf{0.86}(0.00) & 0.26(0.01)\\
    Deep Fair SVDD & 19.76(4.81) & 0.59(0.15) & 0.58(0.22) & 24.65(6.76) & 0.42(0.12) & \textbf{0.71}(0.10)\\
    CFAD & 40.46(2.22) & \textbf{0.74}(0.11) & \textbf{0.78}(0.07) & 32.14(6.57) & 0.38(0.00) & \textbf{0.80}(0.10)\\
    \midrule
    Ex-FairAD (Ours) & \textbf{41.77}(6.28) & 0.54(0.15) & \textbf{0.77}(0.19) & \textbf{34.10}(5.08) & 0.62(0.20) & 0.65(0.10)\\
    Im-FairAD (Ours) & \textbf{43.64}(9.16) & 0.56(0.11) & 0.55(0.26) & \textbf{35.51}(4.11) & \textbf{0.86}(0.00) & 0.62(0.16)\\
        \bottomrule
    \end{tabular}}
\end{table}

\section{Ablation Study}
\label{abalation-study}
In this section, we delve into the influence of different loss terms for our proposed method. More specifically, we investigate the reconstruction error term within the optimization objective of Im-FairAD and the fairness term within the optimization objective of Ex-FairAD.

\subsection{Reconstruction Error Term in Im-FairAD}
\begin{figure}[h!]
    \centering
    \includegraphics[width=0.85\textwidth]{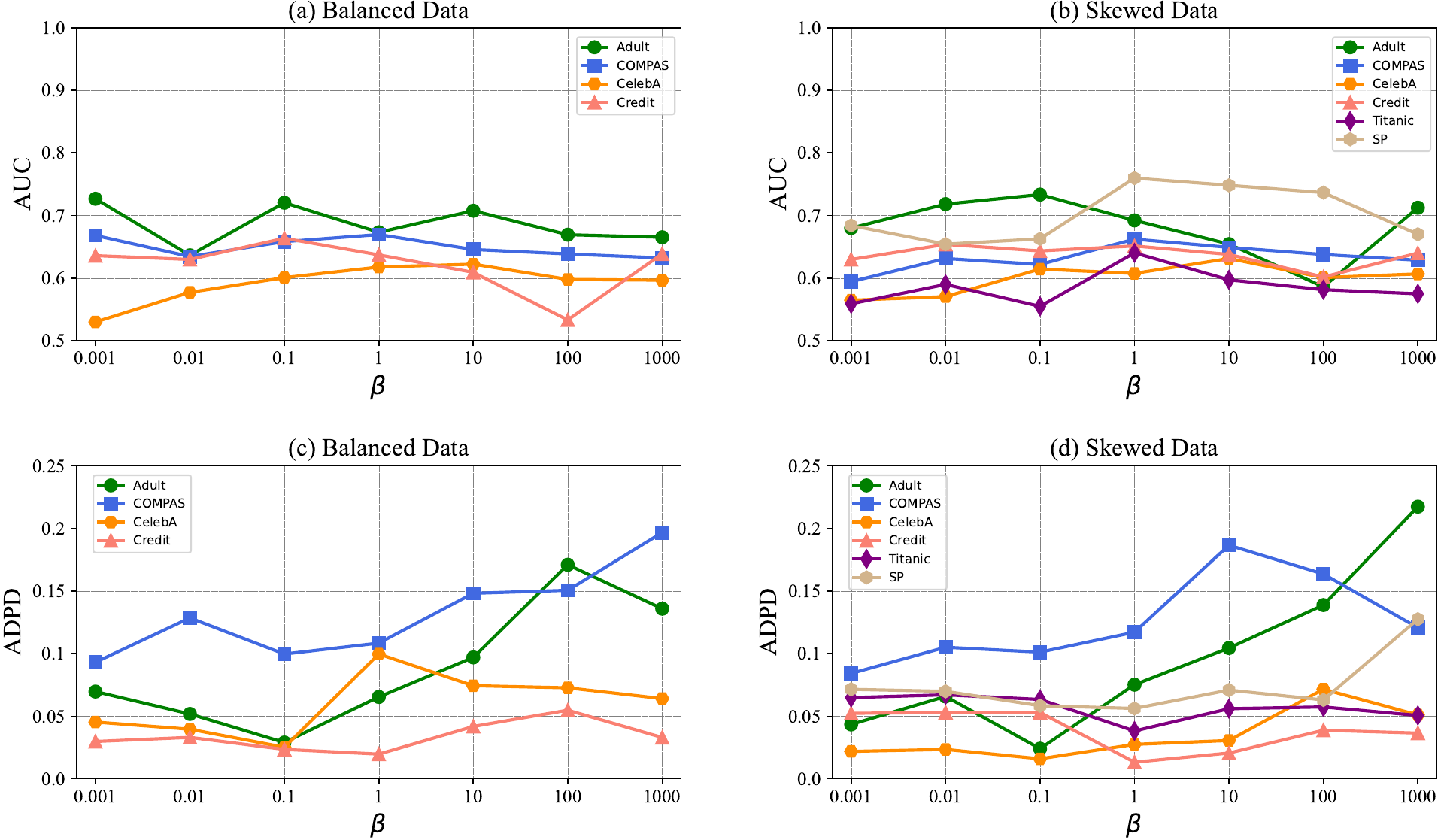}
    \caption{Average AUC and ADPD (all test set) with $\beta$ varies in the range of $\{0.001, 0.01, 0.1, 1, 10, 100, 1000\}$.}
    \label{fig-ablation1}
\end{figure}

For the optimization objective of Im-FairAD, we adjust the hyperparameter $\beta$ across the range $\{0.001, 0.01, 0.1, 1, 10, 100, 1000\}$ to observe the changing of performance, including detection accuracy and fairness. The experimental results are shown in Figure~\ref{fig-ablation1}, where (a) and (b) depict the fluctuation of AUC with varied $\beta$ on balanced and skewed data, respectively. And (c) and (d) depict the fluctuation of ADPD (all  test set) with varied $\beta$ on balanced and skewed data, respectively.
From Figure~\ref{fig-ablation1}, we observe that as $\beta$ increases, the fluctuation of AUC on both balanced and skewed data gradually diminishes. Conversely, the fluctuation of ADPD becomes more pronounced and tends to increase. This observation aligns with expectations, as the dominance of the reconstruction term in the optimization objective makes it challenging to map different protected groups into the same target distribution.

\subsection{Fairness Term in Ex-FairAD}
\begin{figure}[h!]
    \centering
    \includegraphics[width=0.85\textwidth]{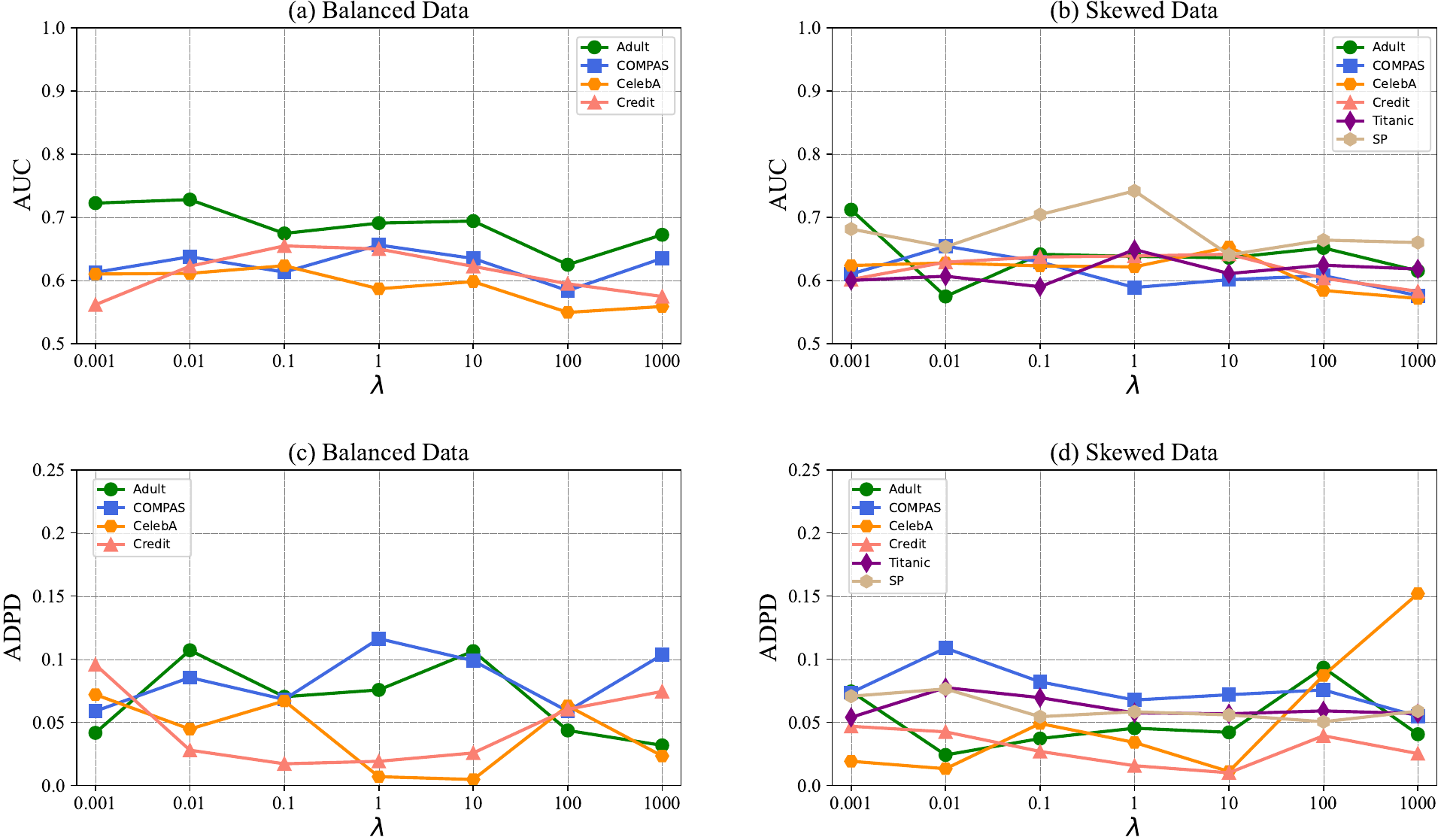}
    \caption{Average AUC and ADPD (all  test set) with $\lambda$ varies in the range of $\{0.001, 0.01, 0.1, 1, 10, 100, 1000\}$ on datasets Adult, COMPAS and CelebA.}
    \label{fig-ablation-lambda}
\end{figure}

For the optimization objective of Ex-FairAD, we remove the fairness regularization term and conduct experiments on all datasets. The results are reported in Table~\ref{tab-ablation2}. Observing Table~\ref{tab-ablation2}, the AUC (detection accuracy) of Ex-FairAD improved but fairness measured by ADPD suffers from adverse effects when without fairness term in optimization objective. Moreover, we adjust the hyperparameter $\lambda$ across the range $\{0.001, 0.01, 0.1, 1, 10, 100, 1000\}$ to observe the changes of performance, including detection accuracy and fairness. The experimental results are visualized in Figure~\ref{fig-ablation-lambda}.

\begin{table}[h]
    \centering
    \caption{Comparison between the objective with fairness regularization and the objective without fairness regularization in optimization problem of Ex-FairAD.}
    \label{tab-ablation2}
    \resizebox{0.85\columnwidth}{!}{
    \begin{tabular}{c|c|c|c|c|c|c|c}
    \toprule
   \multirow{3}{*}{Datasets} & \multirow{4}{*}{Methods} & \multicolumn{3}{c|}{Balanced Split} & \multicolumn{3}{c}{Skewed Split} \\
    \cline{3-8}
    & & \multirow{2}{*}{AUC(\%)~$\uparrow$} & \multicolumn{2}{c|}{ADPD} & \multirow{2}{*}{AUC(\%)~$\uparrow$} & \multicolumn{2}{c}{ADPD} \\
    \cline{4-5} \cline{7-8}
    & & & normal & all & & normal & all \\
    \midrule
     \multirow{2}{*}{COMPAS} &  Ex-FairAD & 63.11 & 4.38 & 4.54 & 66.44 & 5.44 & 7.36\\
     & W/O Fairness term & 67.22 & 13.02 & 12.90 & 68.91 & 14.40 & 12.05\\
    \midrule
    \multirow{2}{*}{Adult} &  Ex-FairAD & 73.49 & 2.92 & 2.86 & 69.56 & 1.52 & 1.79\\
     & W/O Fairness term & 75.32 & 5.56 & 5.39 & 74.94 & 9.29 & 8.07\\
    \midrule
    \multirow{2}{*}{CelebA} & Ex-FairAD & 60.68 & 1.06 & 3.80 & 63.07 & 1.91 & 4.87 \\
     & W/O Fairness term & 62.56 & 3.44 & 5.30 & 64.29 & 3.29 & 3.55\\
    \midrule
    \multirow{2}{*}{Credit} & Ex-FairAD & 64.96 & 2.95 & 1.92 & 63.85 & 2.71 & 1.57 \\
     & W/O Fairness term & 65.40 & 7.21 & 5.87 & 66.63 & 7.33 & 4.64\\
     \midrule
    \multirow{2}{*}{Titanic} & Ex-FairAD & NA & NA & NA & 64.87 & 6.35 & 5.75 \\
     & W/O Fairness term & NA & NA & NA & 68.00 & 10.27 & 9.79\\
     \midrule
    \multirow{2}{*}{SP} & Ex-FairAD & NA & NA & NA & 74.16 & 6.00 & 5.84 \\
     & W/O Fairness term & NA & NA & NA & 76.49 & 8.59 & 8.89\\
    \bottomrule
    \end{tabular}
    }
\end{table}

\section{Selection of Distance Metric $\mathcal{M}$ between Distributions}
\label{com-sinkhorn-mmd}
Based on the analysis on Module Formulation, we obtain the following optimization problem

\begin{equation}\label{optim-1}
    \underset{\phi, \psi}{\text{min}} \sum_{s \in S} \mathcal{M}(\mathcal{D}_{h_\phi(\mathcal{X}_{S=s})}, \mathcal{D}_{\mathbf{z}}) + 
    \beta \mathcal{M}(\mathcal{D}_{g_\psi(h_\phi(\mathbf{x}))}, \mathcal{D}_{\mathbf{x}}).
\end{equation}

However, the problem~\ref{optim-1} is intractable as data distribution $\mathcal{D}_{\mathbf{x}}$ is unknown and $\mathcal{D}_{h_\phi(\mathcal{X}_{S=s})}, \mathcal{D}_{g_\psi(h_\phi(\mathbf{x}))}$ cannot be computed analytically, which leads to that we cannot use f-divergence~\citep{renyi1961measures}, such as KL-divergence, Hellinger distance, to measure the difference between $\mathcal{D}_{h_\phi(\mathcal{X}_{S=s})}$ and $\mathcal{D}_{\mathbf{z}}$. Thus, we need to measure the divergence between $\mathcal{D}_{h_\phi(\mathcal{X}_{S=s})}$ and $\mathcal{D}_{\mathbf{z}}$ by using the known samples. In such situation, Wasserstein distance, Maximum Mean Discrepancy (MMD)~\citep{gretton2012kernel} and Sinkhorn distance~\citep{cuturi2013sinkhorn} are all possible choices.

However, the computation cost of Wasserstein distance can quickly become prohibitive when the data dimension increases. Therefore, the options left are Sinkhorn distance and Maximum Mean Discrepancy.

Based on~\citep{gretton2012kernel}, MMD is defined as
\begin{equation}
    \label{eq3}
          \text{MMD}[\mathcal{F}, p,q] =\underset{\Vert f \Vert_{\mathcal{H}}\leq1}{\sup}\left(\mathbb{E}_p[f(\mathbf{x})]-\mathbb{E}_q[f(\mathbf{y})]\right),
\end{equation}
where $p, q$ are probability distributions, $\mathcal{F}$ is a class of functions $ f:\mathbb{R}^d \rightarrow \mathbb{R}$ and $\mathcal{H}$ denotes a reproducing kernel Hilbert space.

Its empirical estimate is
\begin{equation}
    \label{eq4}
        \begin{aligned}
          &\text{MMD}^2[\mathcal{F}, X,Y] = \frac{1}{m(m-1)}\sum_{i=1}^m\underset{j\neq i}{\sum^m}k(\mathbf{x}_i, \mathbf{x}_j) \\
          &+ \frac{1}{n(n-1)}\sum_{i=1}^n\underset{j\neq i}{\sum^n}k(\mathbf{y}_i, \mathbf{y}_j)
          - \frac{2}{mn}\sum_{i=1}^m\underset{j=1}{\sum^n}k(\mathbf{x}_i, \mathbf{y}_j),
          \end{aligned}
\end{equation}
where $X = \{\mathbf{x}_1, \dots, \mathbf{x}_m\}$ and $Y = \{\mathbf{y}_1, \dots, \mathbf{y}_n\}$ are samples consisting of i.i.d observations drawn from $p$ and $q$, respectively. $k(\cdot, \cdot)$ denotes a kernel function. 

We use Sinkhorn distance and MMD to replace the first term of the optimization problem~\ref{optim-1}, respectively, and use reconstruction error to replace the second term of the problem~\ref{optim-1}. The related experimental results are provided in Table~\ref{tab-sink-mmd}, where we use Gaussian kernel $\exp(-\gamma\|\mathbf{x}-\mathbf{y}\|^2)$ as kernel function of MMD.

\begin{table}[h!]
    \centering
    \caption{The comparison between Sinkhorn distance and MMD on COMPAS.}
    \label{tab-sink-mmd}
    \begin{tabular}{l|c|c|c}
    \toprule
    \multirow{2}{*}{Methods} & \multirow{2}{*}{AUC(\%) $\uparrow$} & \multicolumn{2}{c}{ADPD(\%) $\downarrow$} \\
    \cline{3-4}
    & & normal & all \\
    \midrule
    Im-FairAD (MMD) & 60.78 & 6.75 & 6.81\\
    Im-FairAD (Sinkhorn) & 63.89 & 4.16 & 4.76\\
    \midrule\
    Ex-FairAD (MMD) & 57.09 & 2.60 & 4.43\\
    Ex-FairAD (Sinkhorn) & 63.11 & 4.38 & 4.54\\
    \bottomrule
    \end{tabular}
\end{table}

Based on the empirical results in COMPAS, we select Sinkhorn distance as distribution distance metric of the optimization problem~\ref{optim-1}.

\section{Time Complexity Analysis}
\label{ana-time-complexity}

As most baselines are based on deep learning techniques, the main time cost is from the training of neural networks. Thus, based on our proposed method, we analyze the time complexity of neural networks and other methods have the similar analysis process.

For convenience, the $h_\phi$ is specified as follows
\begin{equation}
h_{\phi}(\mathbf{x}):=\mathbf{W}_{L_h}^h(\sigma(\cdots(\mathbf{W}_2^h(\sigma(\mathbf{W}_{1}^h\mathbf{x})\cdots)),
\end{equation}
where $\phi=\{\mathbf{W}_{1}^h,\mathbf{W}_{2}^h,\ldots,\mathbf{W}_{L_h}^h\}$, $\mathbf{W}_l^h\in\mathbb{R}^{d_l^h\times d_{l-1}^h}$, and $L_h$ is the number of layers of the network. The definition of the layer width indicates that $d_0^h=d$ and $d_{L_h}^h=m$. $\sigma$ denotes the activation function such as ReLU, LeakyReLU, Sigmoid, or Tanh. The activation functions in different layers are not necessarily the same but here we use the same $\sigma$ for convenience. Similarly, the $g_\psi$ is specified as follows
\begin{equation}
g_{\psi}(\mathbf{z}):=\mathbf{W}_{L_g}^g(\sigma(\cdots(\mathbf{W}_2^g(\sigma(\mathbf{W}_{1}^g\mathbf{z})\cdots)),
\end{equation}
where $\psi=\{\mathbf{W}_{1}^g,\mathbf{W}_{2}^g,\ldots,\mathbf{W}_{L_g}^g\}$, $\mathbf{W}_l^g\in\mathbb{R}^{d_l^g\times d_{l-1}^g}$, $d_{1}^g=m$, and $d_{L_g}^g=d$. Note that we have omitted the bias terms of $h_\phi$ and $g_\psi$ for simplicity. 

In the training stage, suppose the batch size of optimization is data size $n$, then the time complexity of training neural network is $\mathcal{O}(T(n\sum_{l=1}^{L_h}d_l^hd_{l-1}^h+n\sum_{l=1}^{L_g}d_l^gd_{l-1}^g))$, where $T$ is the total number of iterations of $h_\phi$ and $g_\psi$. If we further assume $\max(\max_ld^h_{l},\max_ld^g_{l})\leq\bar{d}$, and $L_h+L_g\leq \bar{L}$, the time complexity of neural network is at most $\mathcal{O}(Tn\bar{d}^2\bar{L})$. In addition, for the proposed Im-FairAD and Ex-FairAD, we utilize the Sinkhorn distance to compute the loss. Therefore, the overall time complexity is $\mathcal{O}(T(n\bar{d}^2\bar{L} + tn^2))$ where the time complexity of Sinkhorn algorithm is $\mathcal{O}(n^2)$ and $t$ is the maximum iterations of Sinkhorn algorithm.

In the inference phase, for $m$ new samples, the time complexity of computing the anomaly score is $\mathcal{O}(m\bar{d}^2\bar{L} + m)$ where $\mathcal{O}(m\bar{d}^2\bar{L})$ is from neural networks and $\mathcal{O}(m)$ is from the calculation of anomaly score. The detailed comparison of time complexity is provided in Table~\ref{tab-time-complexity}, where all the methods including ours have the same time complexity in inference phase.
\begin{table}[h!] 
    \centering
    \caption{The time complexity of training and inference.} \label{tab-time-complexity}
    \resizebox{0.85\columnwidth}{!}{
    \begin{tabular}{l|c|c}
    \toprule
    Methods & Time Complexity (Training) & Time Complexity (Inference) \\
    \midrule
    Deep SVDD & $\mathcal{O}((T_{ae} + T_{oc})(n\bar{d}^2\bar{L} + n))$ & $\mathcal{O}(m\Bar{d}^2\bar{L} + m)$\\
    FairOD & $\mathcal{O}(T(n\bar{d}^2\bar{L} + n))$ & $\mathcal{O}(m\Bar{d}^2\bar{L} + m)$\\
    Deep Fair SVDD & $\mathcal{O}((T_{ae} + T_d + T_g)(n\bar{d}^2\bar{L} + n))$ & $\mathcal{O}(m\Bar{d}^2\bar{L} + m)$\\
    CFAD & $\mathcal{O}((T_{gae} + T_d + T_{ae} + T_c)(n\bar{d}^2\bar{L} + n))$ & $\mathcal{O}(m\Bar{d}^2\bar{L} + m)$\\
    \midrule
    Ex-FairAD & $\mathcal{O}(T(n\Bar{d}^2\bar{L} + tn^2 ))$ & $\mathcal{O}(m\Bar{d}^2\bar{L} + m)$\\
    Im-FairAD & $\mathcal{O}(T(n\Bar{d}^2\bar{L} + tn^2 ))$ & $\mathcal{O}(m\Bar{d}^2\bar{L} + m)$\\
    \bottomrule
    \end{tabular}
    }
\end{table}

\section{The Effectiveness of Equal Opportunity}
\label{eo-effectiveness}
According to the definition~\eqref{def_EO_mc} of equal opportunity, we define EO as a fairness metric that can measure equal opportunity:
\begin{equation}\label{def_eom}
\begin{aligned}
        \text{EO} := &\vert \mathbb{P}[\hat{y}=1~|~S=s_i, y=1] -  \mathbb{P}[\hat{y}=1~|~S=s_j, y=1]\vert\\
        = &\vert\mathbb{P}[\mathcal{T}^*(\mathbf{x}) > t~|~\mathbf{x} \in \mathcal{X}_{S=s_i}, y=1] \\
        - &\mathbb{P}[\mathcal{T}^*(\mathbf{x}) > t~|~\mathbf{x} \in \mathcal{X}_{S=s_j}, y=1]\vert
\end{aligned}
\end{equation}
In this section, we explore the effectiveness of equal opportunity in unsupervised anomaly detection. According to definition~\ref{def_EO_mc} of equal opportunity, we need to measure the fairness on abnormal data. Therefore, we calculate the ADPD and EO~\eqref{def_eom} only on abnormal data. To determine threshold for calculating EO, we sort the anomaly scores of the training set in ascending order. The threshold $t$ is then set to the $pN$-th smallest anomaly score, with $p$ fixed at $0.95$ for all methods, and $N$ denoting the size of the training set. we report the experimental results in Table~\ref{tab-compas-EQ} and Table~\ref{tab-adult-EQ}. From the Table~\ref{tab-compas-EQ} and Table~\ref{tab-adult-EQ}, we have the following observations:

\begin{table}[h!] 
    \centering
    \caption{Detection accuracy (AUC) and fairness (ADPD and EO) on COMPAS. Note that ADPD is measured only on abnormal data, which makes it comparable with EO.}
    \label{tab-compas-EQ}
    \resizebox{0.85\textwidth}{!}{
    \begin{tabular}{l|c|c|c|c|c|c}
    \toprule
    \multirow{3}{*}{Methods} & \multicolumn{3}{c|}{Balanced Split} & \multicolumn{3}{c}{skewed Split} \\
    \cline{2-7}
    & \multirow{2}{*}{AUC~$\uparrow$} & \multicolumn{2}{c|}{Fairness ~$\downarrow$} & \multirow{2}{*}{AUC~$\uparrow$} & \multicolumn{2}{c}{Fairness~$\downarrow$} \\
    \cline{3-4}\cline{6-7}
    & & ADPD(\%) & EO(\%) & & ADPD(\%) & EO(\%) \\
    \midrule
    LOF & 59.23 & 12.99 & 14.82 &  57.25 & 12.91 & 12.47\\
    Deep SVDD & 57.58 & 18.97 & 8.99 & 60.24 & 23.82 & 12.54 \\
    FarconVAE+LOF & 48.72 & 6.13 & 3.37 & 50.10 & \textbf{4.84}& \textbf{2.41} \\
    FarconVAE+Deep SVDD & 50.48 & 6.69 & \textbf{1.54} & 48.83 & 6.02 & \textbf{3.23}\\
    
    FairOD & 61.05 & 6.51 & \textbf{1.49} & 58.86 & 10.87 & 5.82\\
    
    Deep Fair SVDD & 62.71 & 7.46 & 8.99 & 61.05 & 11.61 & 11.74\\
    
    CFAD & 62.27 & 10.55 & 14.63 & 60.87 & 16.50 & 6.69 \\
    \midrule
    Ex-FairAD (Ours) & \textbf{63.11} & \textbf{5.56} & 4.35 & \textbf{66.44} & \textbf{4.88} & 5.34\\
    
    Im-FairAD (Ours) & \textbf{63.89} & \textbf{4.70} & 4.13 & \textbf{66.58} & 8.95 & 5.09\\
    \bottomrule
    \end{tabular}}
\end{table}

\begin{table}[h!]
    \centering
    \caption{Detection accuracy (AUC) and fairness metrics (ADPD and EO) on Adult. Note that ADPD is measured only on abnormal data, which makes it comparable with EO.}
    \label{tab-adult-EQ}
    \resizebox{0.85\textwidth}{!}{
    \begin{tabular}{l|c|c|c|c|c|c}
    \toprule
    \multirow{3}{*}{Methods} & \multicolumn{3}{c|}{Balanced Split} & \multicolumn{3}{c}{skewed Split} \\
    \cline{2-7}
    & \multirow{2}{*}{AUC~$\uparrow$} & \multicolumn{2}{c|}{Fairness ~$\downarrow$} & \multirow{2}{*}{AUC~$\uparrow$} & \multicolumn{2}{c}{Fairness~$\downarrow$} \\
    \cline{3-4}\cline{6-7}
    & & ADPD(\%) & EO(\%) & & ADPD(\%) & EO(\%) \\
    \midrule
    LOF & 58.09 & 26.38 & 2.89 & 59.14 & 28.50 & 27.05\\
    Deep SVDD & 60.47 & 10.95 & 6.85 & 61.04 & 21.81 & 5.92\\
    FarconVAE+LOF & 53.92 & 7.89 & \textbf{2.20} & 50.56 & 8.09 & 4.85\\
    FarconVAE+Deep SVDD & 55.05 & 8.28 & 6.68 & 51.21 & 7.26 & 9.64\\
    
    FairOD & 70.63 & 13.57 & 15.29 & 58.30 & 27.13 & 31.29\\
    
    Deep Fair SVDD & 62.45 & 7.24 & 8.63 & 60.22 & 9.35 & \textbf{3.89}\\
    
    CFAD & 66.89 & 22.38 & 42.29 & 60.28 & 36.41 & 59.89 \\
    \midrule
    Ex-FairAD (Ours) & \textbf{73.49} & \textbf{3.78} & 6.33 & \textbf{69.56} & \textbf{4.97} & 8.74\\
    
    Im-FairAD (Ours) & \textbf{72.05} & \textbf{5.85} & \textbf{2.73} & \textbf{69.34} & \textbf{3.60} & \textbf{4.38}\\
    \bottomrule
    \end{tabular}}
\end{table}
\begin{itemize}
    \item In the transition from a balanced split to a skewed split, it can be observed that equal opportunity, as measured by ADPD and EO, exhibits significantly larger values in most cases which means a skewed split tends to introduce more unfairness compared to a balanced split, which is consistent with the observation from results on normal and overall test set. This indicates that a skewed split poses a more intractable problem for fairness than a balanced split.
    \item FairOD demonstrate superior equal opportunity on the COMPAS dataset compared to Deep SVDD, a fairness-unaware AD method. This observation suggests that unsupervised anomaly detection methods have the potential to ensure equal opportunity to some extents, especially when guided by reasonable assumptions (e.g., Assumption 2 proposed in this paper). However, on the Adult dataset, FairOD and CFAD exhibit poorer equal opportunity than Deep SVDD in both balanced and skewed splits, which indicates that existing fairness-aware unsupervised AD methods are unable to maintain equal opportunity effectively across different data domain.
    \item Compared to all baselines, our methods (Ex-FairAD and Im-FairAD) achieve better detection accuracy (AUC) while maintaining comparable or even better equal opportunity (ADPD and EO) in most cases. This observation supports the reasonability and practicality of Assumption 2 for our methods.
\end{itemize}

\section{Experiments on Contaminated Training Set}
\label{exp-contaminated}
In our main experiments, all methods including the baselines and ours focus on the standard setting of unsupervised anomaly detection~\citep{Ruff_Vandermeulen_Goernitz_Deecke_Siddiqui_Binder_Müller_Kloft_2018,cai2022perturbation,fu2024dense}, that is the training set consists of only normal samples. However, in real scenarios, a small number of unknown abnormal samples may be mixed in the training set. Based on this consideration, we added $1\% (\text{abnormal}/\text{normal})$ anomalous samples to the balanced training set and keep the test set unchanged. The related results are provided in Table~\ref{tab-contamination}, where the detection accuracy of almost all methods has a slight decrease and our proposed methods still achieve better or comparable detection accuracy and fairness in comparison to all baselines.

\begin{table}[h!]
    \centering
    \caption{The results on contaminated training set of COMPAS and Adult.}
    \label{tab-contamination}
    \resizebox{0.85\textwidth}{!}{
    \begin{tabular}{l|c|c|c|c|c|c}
    \toprule
    \multirow{3}{*}{Methods} & \multicolumn{3}{c|}{COMPAS} & \multicolumn{3}{c}{Adult} \\
    \cmidrule{2-7}
    & \multirow{2}{*}{AUC(\%)~$\uparrow$} & \multicolumn{2}{c|}{APDP(\%)~$\downarrow$} & \multirow{2}{*}{AUC(\%)~$\uparrow$} & \multicolumn{2}{c}{APDP(\%)~$\downarrow$} \\
    \cmidrule{3-4} \cmidrule{6-7}
    & & normal & all & & normal & all  \\
    \midrule
    LOF & 54.70 & 5.28 & 5.79 & 49.83 & \textbf{1.07} & 2.01\\
    Deep SVDD & \textbf{62.41} & 26.91 & 18.23 & 62.70 & 5.69 & 7.53\\
    FarconVAE+LOF & 46.15 & 4.53 & \textbf{1.78} & 51.58 & 3.43 & 2.28\\
    FarconVAE+Deep SVDD & 52.82 & 2.73 & \textbf{1.83} & 51.36 & 2.16 & 2.54\\
    FairOD & 54.43 & \textbf{2.20} & \textbf{3.07} & 69.76 & 9.28 & 8.85\\
    Deep Fair SVDD & 60.09 & 9.50 & 8.65 & 62.02 & \textbf{2.07} & 3.74\\
    CFAD & 60.81 & 11.43 & 9.05 & 67.98 & 6.80 & 15.44\\
    \midrule
    Ex-FairAD (Ours) & \textbf{61.67} & 2.82 & 5.66 & \textbf{73.10} & 4.77 & \textbf{1.76}\\
    Im-FairAD (Ours) & 61.10 & \textbf{2.38} & 4.95 & \textbf{71.80} & 5.20 & \textbf{1.31}\\
    \bottomrule
    \end{tabular}
    }
\end{table}

\end{document}